\documentclass[referee,pdflatex,sn-vancouver-num]{sn-jnl}

\usepackage{graphicx}
\usepackage{multirow}
\usepackage{amsmath,amssymb,amsfonts}
\usepackage{amsthm}
\usepackage{xcolor}
\usepackage{booktabs}
\usepackage{algorithm}
\usepackage{algpseudocode}
\usepackage[export]{adjustbox}
\usepackage{subcaption}
\usepackage{tikz}
\usetikzlibrary{fit,positioning}
\usepackage{siunitx}
\usepackage{bbm,accents}

\graphicspath{{./}}

\newcommand{\mb}{\mathbf}

\newcommand{\mc}{\mathcal}

\newcommand{\bb}{\mathbb}

\newcommand{\abs}[1]{\left| #1 \right|}
\newcommand{\innerprod}[2]{\left\langle #1,  #2 \right\rangle}

\newcommand{\expect}[2][]{\bb E_{#1}\left[#2 \right]}

\renewcommand{\mathbf}{\boldsymbol}

\DeclareMathOperator{\trace}{tr}

\DeclareMathOperator{\diag}{diag}

\DeclareMathOperator{\iid}{\stackrel{\text{i.i.d}}{\sim}}

\DeclareRobustCommand{\rev}[1]{#1}
\DeclareRobustCommand{\revii}[1]{#1}

\begin{document}
	
	
	
	\title[Hierarchical sparse Bayesian multitask model]{Hierarchical Sparse Bayesian Multitask Learning for Disease Prediction in Pooled Microbiome Studies}

	\author[1,2]{\fnm{Haonan} \sur{Zhu}}
	\author*[1]{\fnm{Andre R.} \sur{Goncalves}}\email{goncalves1@llnl.gov}
	\author[1]{\fnm{Car Reen} \sur{Kok}}
	\author[1]{\fnm{Camilo} \sur{Valdes}}
	\author[1]{\fnm{Hiranmayi} \sur{Ranganathan}}
	\author[1]{\fnm{Boya} \sur{Zhang}}
	\author[1]{\fnm{Jose Manuel} \sur{Mart\'i}}
	\author[1]{\fnm{Monica K.} \sur{Borucki}}
	\author[1]{\fnm{Nisha J.} \sur{Mulakken}}
	\author[1]{\fnm{James B.} \sur{Thissen}}
	\author[1]{\fnm{Crystal} \sur{Jaing}}
	\author[2]{\fnm{Alfred} \sur{Hero}}
	\author*[1]{\fnm{Nicholas A.} \sur{Be}}\email{be1@llnl.gov}
	
	\affil[1]{\orgname{Lawrence Livermore National Laboratory}, \orgaddress{\city{Livermore}, \state{CA}, \country{USA}}}
	\affil[2]{\orgdiv{Electrical and Computer Engineering}, \orgname{University of Michigan}, \orgaddress{\city{Ann Arbor}, \state{MI}, \country{USA}}}
	
	\abstract{
		\textbf{Background:}
		Microbiome-based disease prediction across pooled studies is challenging because the data are high dimensional, heterogeneous, and often too limited to support reliable study-specific models. We propose a hierarchical sparse Bayesian multitask logistic regression model that encourages shared sparsity across related studies while retaining interpretability and uncertainty quantification. To make posterior inference scalable, we derive a variational approximation for the model parameters.
		
		\textbf{Results:}
		We evaluate the method on synthetic datasets and on a pooled metagenomic collection comprising 61 previously published microbiome studies spanning 19 disease conditions. In simulation, the proposed approach improves support recovery when regression coefficients share a common sparse structure across tasks. \rev{On the pooled microbiome application, the proposed method achieves competitive predictive performance while offering interpretability, consistently identifying sparse sets of informative taxa and providing posterior uncertainty summaries and calibration diagnostics under substantial cross-study heterogeneity.}
		
		\textbf{Conclusion:}
		These results suggest that the method is particularly useful as an interpretable, uncertainty-aware framework for extracting shared microbial signals from heterogeneous microbiome studies.
	}
	
	
	
	\maketitle
	
	\section{Background}
	\label{sec: intro}
	
	The human microbiome has an immense impact on human health, and a growing body of work demonstrates its promise as a basis for disease prediction, including in gastrointestinal (GI) health \cite{lee2021inflammatory,nishida2018gut,mcgrattan2019diet,de2019microbiome,thaiss2016microbiome}, cancer \cite{banerjee2017microbial,sepich2021microbiome,gopalakrishnan2018influence}, neurological disease \cite{boddy2021gut,mcgrattan2019diet}, and immune dysfunction \cite{de2019microbiome,thaiss2016microbiome}. This has motivated increasing interest in employing machine learning models to perform health prediction based on human gut microbiome data \cite{zhou2019review,aryal2020machine,cammarota2020gut,manandhar2021gut}. In practice, however, microbiome data poses particular difficulties for standard machine learning approaches. First, microbiome datasets are high dimensional and small in sample size, making overparameterized models prone to overfitting. Second, clinical applications require interpretable models that quantify prediction uncertainty, a property most deep neural networks lack \cite{molnar2020interpretive}. One natural response to the sample-size problem is pooling data across studies, but this introduces its own complications.
	
	While pooling microbiome studies offers an opportunity to improve disease prediction and feature discovery, it introduces substantial heterogeneity arising from differences in cohort composition, experimental design, laboratory protocols, and sequencing pipelines \cite{goncalves2026}. This heterogeneity makes it difficult to obtain models that are both stable and interpretable, a particularly important property in bioinformatics, where investigators are often interested not only in predictive accuracy but also in identifying microbial features that can be validated experimentally. \rev{Accordingly, the objective of BayesMTL is not to remove all cross-study batch effects, but to model each study as a related task so that task-specific effects can be estimated while a shared sparse microbial signal is learned across tasks.}
	
	In this work, we propose a hierarchical sparse Bayesian multitask logistic regression model for binary classification across related microbiome studies. The model uses a shared sparsity structure to identify taxa that are informative across tasks while allowing task-specific effect sizes. Relative to optimization-based multitask methods \cite{liu2009multi,jalali2010dirty,gonccalves2016multi}, the Bayesian formulation provides posterior uncertainty estimates and a principled way to encode prior assumptions about relationships among studies.
	
	Our approach also differs from prior multitask sparse models \cite{xue2007multi,zhang2010learning,liu2009multi,jalali2010dirty,guo2011sparse,goncalves2019bayesian} in how the sparsity pattern is represented. Following \cite{goncalves2019bayesian}, we use an overparameterized Bernoulli-Gaussian formulation rather than regularization penalties, which has been associated with improved support recovery \cite{bazot2011bernoulli}. Because the posterior distribution is not available in closed form, we use variational inference to obtain a scalable approximation for high-dimensional microbiome feature spaces.
	
	We evaluate the model on synthetic datasets and on a pooled metagenomic resource containing $61$ previously published studies across $19$ disease conditions. The simulation experiments show that the approach performs well when the underlying tasks share sparse structure, particularly in support recovery. \rev{On the real microbiome data, the main benefit is not uniform improvement in predictive accuracy, but rather competitive prediction together with sparse, interpretable microbial signatures, posterior uncertainty summaries, and calibration diagnostics that reveal overconfidence at probability extremes.}

	\section{Methods}
	
	\subsection{Data collection and preprocessing}

	To ensure comparability across studies, all raw DNA sequence data were processed using a unified pipeline. Sequence data were downloaded from the appropriate repository for each of the 61 studies referenced above. All datasets consisted of short-read data generated on Illumina sequencing platforms. Reads were quality filtered with fastp \cite{chen2018fastp}. Metagenomic classification and taxonomic assignment were performed with Centrifuge \cite{kim2016centrifuge,marti2024addressing}, using a reference index built from the NCBI Nucleotide (nt) database \cite{ncbi2012database} as of January 4, 2023. Post-processing and filtering with a minimum hit length of 40 were performed using Recentrifuge \cite{marti2019recentrifuge}. The output consisted of read counts assigned to each observed taxonomic identifier (taxID) in the reference index. Full details of the data collection and processing pipeline, including the complete list of studies, are provided in \cite{kok2024meta2db}. The list of all microbiome studies used in this work and their associated metadata is provided in Section 1 of the Supplementary Information.
	
	\rev{Using a unified preprocessing workflow reduces technical variation due to differences in read filtering, taxonomic assignment, and feature construction, but it does not eliminate residual batch effects. Residual technical or cohort-level effects can still influence prediction and feature selection, especially when they are confounded with disease status, geography, body site, or cohort composition. BayesMTL should therefore be interpreted as a multitask model that accommodates study-specific coefficients rather than as an explicit batch-correction procedure.} \revii{We fit BayesMTL, and all baseline methods, directly to these features without an upstream batch-effect correction. BayesMTL is nonetheless compatible with such a step: an appropriate method can be applied to the features before fitting. Because study and disease status are partially confounded in this pooled design, such a correction should be chosen to preserve disease-associated variation.}
	
	Read count values for each taxonomic identifier were further processed by centered log ratio (CLR) transform \cite{aitchison1982statistical} to provide features for the machine learning model. Seven total taxonomic rank levels were examined for purposes of this analysis, and we take the union set of all the taxIDs for each taxonomic rank to provide a consistent feature space. \rev{Thus, studies with partially overlapping observed taxa are aligned to a common rank-specific taxID dictionary. A taxID that is not observed in a study is represented by the corresponding absent count in that study before feature transformation, so identical observed feature sets are not required across studies.} The taxonomic ranks are: Kingdom ($11$), Phylum ($160$), Class ($313$), Order ($868$), Family ($2,338$), Genus ($7,519$) and Species ($31,679$), where each number in the parentheses corresponds to the feature dimension (i.e., total number of taxIDs). The health condition studied for each patient is one-hot encoded into one of the $19$ conditions (see Supplementary Information document).

	\subsection{Notations}
	\label{subsec: notations}
	
	We refer to the abundance of each taxonomic identifier (taxID) as a feature. Each study is referred to as a task, and the disease status of each individual is treated as a binary label. Bold upper-case letters denote matrices, bold lower case letters denote vectors and no bold lower case denote scalars. The Hadamard (element-wise) product of vectors $\mb a$ and  $\mb b$ is denoted by  $\mb a \circ \mb b$, and $\diag$ denotes the function map a vector to a diagonal matrix with the vector as its diagonal entries. We denote the observed taxID count data after centered log ratio transform  as $\mb x_{t}^{i} \in \bb R^{d}$ together with its label $y_{t}^{i} \in \{0,1\}$, where $d$ corresponds to the number of features (i.e., number of taxIDs),  $t=1,\ldots, T$ denotes the different tasks (i.e., different studies), $i=1,\ldots, n_t$ denotes the different subjects per study, $n_{t}$ denotes the total number of subjects per task, and $y_{t}^{i}$ reflect whether the subject is diseased ($1$) or control ($0$). For a given regression weight matrix $\mb W \in \bb R^{T\times d}$ for all the tasks, we denote $\mb w_{t} \in \bb R^{d}$ the row vector of $\mb W$  corresponds to $t$-th task across features, and $\mb w_{\left( j \right) } \in \bb R^{T}$ the column vector of $\mb W$ corresponds to $j$-th feature across tasks.

	\subsection{Hierarchical sparse Bayesian multitask logistic regression model}
	\label{subsec: bmtl model}
	
	Although the human microbiome comprises a vast number of microbial species, only a small subset is typically associated with disease \cite{wang2017human}, motivating the assumption that only a sparse subset of taxa is informative for disease prediction. We encode this assumption with a Bernoulli-Gaussian prior \cite{soussen2011bernoulli,bazot2011bernoulli} in which a binary inclusion vector $\mb z$ determines whether a feature is active and the continuous weights $\mb w_t$ determine the effect size for each task. Because $\mb z$ is shared across tasks, the model favors a common sparse support, while still allowing the magnitude of each effect to vary from study to study. \rev{This construction lies between complete pooling and independent single-task fitting: the support variables are shared, but the coefficients remain task-specific, so a selected taxon can have different magnitudes or signs across studies and can be effectively negligible in a subset of tasks.} Excluded taxa are assigned exactly zero weight rather than merely shrunk, and a beta hyperprior on the inclusion probability allows the overall sparsity level to be inferred from the data.
	
\begin{figure*}[t]
		\centering
		\begin{subfigure}{0.58\textwidth}
			\centering
			\includegraphics[width=\textwidth]{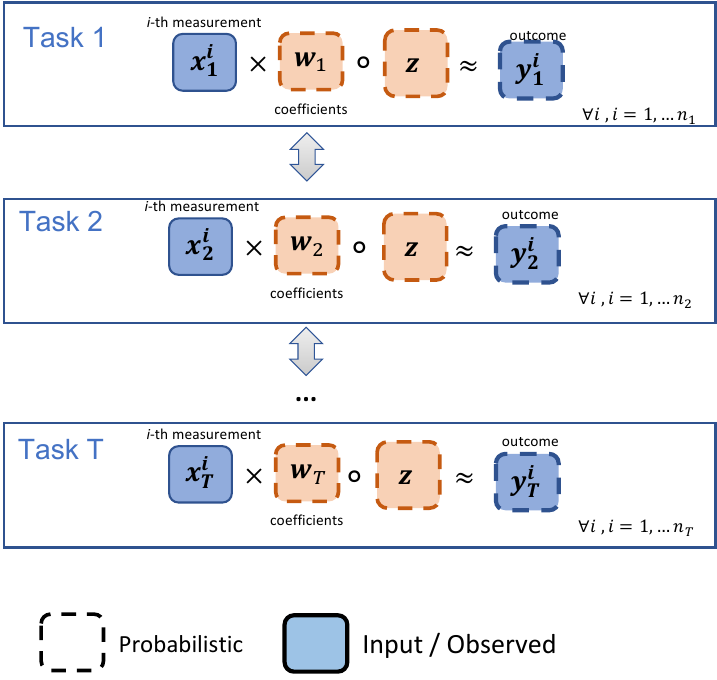}
			\caption{The based model}
			\label{subfig: base model}
		\end{subfigure}
		\begin{subfigure}{0.58\textwidth}
			\centering
			\includegraphics[width=\textwidth]{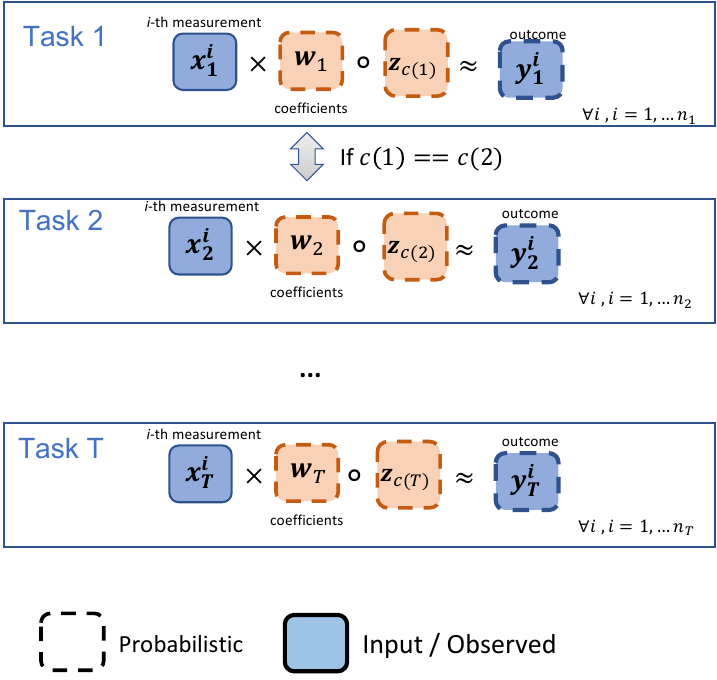}
			\caption{The clustering extension}
			\label{subfig: cluster extension}
		\end{subfigure}
		\caption{Graphical visualization of the proposed probabilistic model. For the base model in (a), the information sharing cross different tasks is enforced by common priors between the regression weights and sparsity parameters. For the clustering extension (b), the information sharing is limited to tasks from the same cluster (i.e., disease category).}
		\label{fig: model}
	\end{figure*}

	Further, to enforce the information sharing across different tasks, the row of $\mb W$, denoted by $\mb w_{\left( j \right)}$ are assumed to be $i.i.d$ draws from a multivariate Gaussian distribution with mean $\mb 0$ and covariance $\mb \Sigma_0$. A Wishart prior is proposed for this shared covariance matrix to provide a method for utilizing expert knowledge about the underling relationships among the studies. This enables us to exploit the Wishart-normal conjugacy to obtain efficient inference later. The proposed conditional model can be summarized:
	
	{\small \begin{align*}
			y_{t}^{i}|\mb w_{t}, \mb z; \mb x_{t}^{i} &\stackrel{\text{independent}}{\sim} \text{Bernoulli}\left( \text{sigmoid}\left( \innerprod{\left(\mb w_{t}\circ \mb z\right)}{\mb x_{t}^{i}} \right)   \right) \\
			& \qquad \forall i=1,\ldots, n_{t},\forall t=1,\ldots, T, \\ 
			z_{j}|\theta & \iid \text{Bernoulli} \left( \theta \right) \quad \forall j=1,\ldots, d,\\  
			\theta &\sim \text{Beta}\left( \alpha_0,\beta_0 \right), \\ 
			\mb w_{\left( j \right) }|\mb \Sigma_0 & \iid \mc N\left( \mb 0, \mb \Sigma_0 \right) \quad \forall j=1,\ldots, d,\\
			\mb \Sigma_0^{-1} & \sim \text{Wishart}\left( v_0,\mb V_0 \right).  
	\end{align*}}
	
	\noindent where $\alpha_0,\beta_0, v_0, \mb V_0$ are hyperparameters. \rev{The beta prior encodes the expected prior inclusion probability as $\alpha_0/(\alpha_0+\beta_0)$, while the concentration $\alpha_0+\beta_0$ controls the strength of this sparsity prior. Smaller values of $\alpha_0/(\alpha_0+\beta_0)$ therefore express a prior belief that fewer features are informative. The Wishart parameters $v_0$ and $\mb V_0$ encode prior information about the cross-task covariance structure of the regression coefficients. In the experiments, these hyperparameters are selected on the training data by the repeated stratified validation procedure described in Section~\ref{subsec: experimental setup} and detailed in the Supplementary Information section titled ``Experimental setup and reproducibility details''.} See Fig. \ref{fig: model} for a visualization of the proposed model and its cluster extension used in subsection \ref{subsec: microbiome data}.
	
	The model induces the following conditional log-probability, up to an additive constant, where $\Gamma$ denotes the gamma function.
	
	{
		\small
		\begin{align}
			\label{eqn: cond log-likelihood}
			&\log\left( p\left( \mb Y, \mb W, \mb \Sigma_0, \theta, \mb z \mid  \mb X; v_0, \mb V_0, \alpha_0, \beta_0 \right)  \right) \nonumber \\
			&= -\frac{1}{2}\trace\left( \mb V_0^{-1} \mb \Sigma_0^{-1} \right) \nonumber \\
			&\quad +\frac{v_0+d-T-1}{2}\log\det\left( \mb \Sigma_0^{-1} \right) \nonumber \\
			&\quad -\frac{v_0}{2}\log\det\left( \mb V_0 \right) \nonumber \\
			&\quad +\left( \alpha_0-1 \right) \log\left( \theta \right) +\left( \beta_0-1 \right) \log\left( 1-\theta \right) \nonumber \\
			&\quad +\log\Gamma\left( \alpha_0+\beta_0 \right)-\log\Gamma\left( \alpha_0 \right)-\log\Gamma\left( \beta_0 \right) \nonumber \\
			&\quad -\sum_{j} \frac{1}{2} \innerprod{\mb w_{\left( j \right) }}{\mb \Sigma_0^{-1} \mb w_{\left( j \right) }} \nonumber \\
			&\quad + \sum_{t}\sum_{i} y_{t}^{i} \left( \innerprod{\left(\mb w_{t} \circ \mb z\right)}{\mb x_{t}^{i}} \right)  \nonumber \\
			&\quad -\sum_{t}\sum_{i} \log\left( \exp\left( \innerprod{\left( \mb w_{t}\circ \mb z\right)}{\mb x_{t}^{i}}\right)+1\right) \nonumber \\
			&\quad + \left( \sum_{j} z_{j} \right) \log \theta + \left( d-\sum_{j} z_{j} \right) \log\left( 1-\theta \right).
		\end{align}
	}

	\subsection{Variational inference}
	\label{sec: bmtl Inference}
	
	With the combination of the logistic function and the hierarchical structure, inference from the exact posterior distribution of the conditional model is intractable since the posterior distribution is not available in closed form. We resort to a variational approach \cite{blei2017variational,zhang2018advances} where we approximate the posterior distribution with a simpler distribution, and the approximation is iteratively refined. We refer interested readers to \cite{blei2017variational} for a comprehensive review on variational inference (VI) as a general approach for Bayesian inference. 
	
	Section \ref{subsec: mean field approximation and VI bound} introduces the mean-field approximation used to approximate the posterior distribution along with the variational objective function, and Section \ref{subsec: CAVI} summarizes the optimization algorithm based on coordinate ascent \cite{bishop2006pattern}.
	
	\subsubsection{Mean-field approximation and variational lower bound}
	\label{subsec: mean field approximation and VI bound}
	
	We adopt a mean-field variational family because it balances flexibility and computational tractability. The approximation factorizes the posterior over the global sparsity parameter, the shared covariance matrix, the binary inclusion variables, and the task-specific regression coefficients. This factorization makes it possible to derive closed-form updates for most terms while preserving the key latent structure of the model.

	In this work, we propose to approximate the true posterior of the proposed model by the function:
	\begin{align}
		\label{eqn: MF-approximation}
		q( \theta, \mb W, \mb \Sigma_0, \mb z) &= q\left( \theta; \alpha,\beta \right)  q\left( \mb \Sigma_0^{-1}; v, \mb V \right) \nonumber \\ 
		& \prod_{j} q\left( z_{j};\phi_{j} \right) q( \mb w_{\left( j \right) }; \mb m_{\left( j \right) }, \mb \Sigma_{j} ).
	\end{align}
	
	Here, $q\left( \theta;\alpha,\beta \right)$ is a beta distribution with parameters $\left(\alpha,\beta \right) $, $q\left( z_{j};\phi_{j} \right)$ is a Bernoulli distribution with parameters $\phi_{j}$, $q\left( \mb \Sigma_0^{-1};v, \mb V \right)$ is a Wishart distribution with parameters $\left( v, \mb V \right)$, and $q\left( \mb w_{\left( j \right)}; \mb m_{\left( j \right)}, \mb \Sigma_{j} \right)$ is a multivariate Gaussian distribution with mean $\mb m_{\left( j \right) }$ and covariance $\mb \Sigma_{j}$ for $j=1,\ldots, d$. Together, these factors define the variational family used to approximate the true posterior.
	
	The optimization objective of variational inference (VI) is to minimize the KL-divergence between the approximation Eqn. \ref{eqn: MF-approximation} and the true posterior by maximizing the evidence lower bound (ELBO): 
	{\small 
		\begin{align}
			\label{eqn: elbo}
			\text{ELBO}(q) &= \expect[q]{\log f(\mb X, \mb Y, \theta, \mb z, \mb W, \mb \Sigma_0)} + \text{Entropy}(q) \nonumber \\
			&= -\frac{1}{2} \trace (v \mb V_{0}^{-1} \mb V) + \frac{v_0+d}{2} \log\det (\mb V) \nonumber \\
			&\quad + \frac{v_0+d-v}{2} \psi_{T} \left( \frac{v}{2} \right) + \frac{vT}{2}+\ln \Gamma_{T} \left( \frac{v}{2} \right) \nonumber \\
			&\quad + \left( \alpha_0+\sum_{j} \phi_{j}-\alpha \right) \psi(\alpha) +\ln B(\alpha,\beta) \nonumber \\
			&\quad +\left( \beta_0+d-\sum_{j} \phi_{j} -\beta \right)\psi(\beta) \nonumber \\
			&\quad + \left( \alpha+\beta-d-\alpha_0-\beta_0 \right) \psi(\alpha+\beta) \nonumber \\
			&\quad -\frac{1}{2} \sum_{j} \trace (v \mb V (\mb m_{(j)} \mb m_{(j)}^{\top}+\mb \Sigma_{j})) \nonumber \\
			&\quad +\sum_{t}\sum_{i} y_{t}^{i} \left( \innerprod{(\mb m_{t}\circ \mb \phi)}{\mb x_{t}^{i}} \right) \nonumber \\ 
			&\quad - \sum_{t}\sum_{i}\expect[\mb W, \mb z\sim q]{\log\left( \exp\left( \innerprod{(\mb w_{t}\circ \mb z)}{\mb x_{t}^{i}} \right)+1 \right)} \nonumber \\
			&\quad +\frac{1}{2}\sum_{j}\log\det (\mb \Sigma_{j})- \sum_{j} \phi_{j} \log(\phi_{j}) \nonumber\\ 
			&\quad - \sum_{j} (1-\phi_{j})  \log(1-\phi_{j}).
		\end{align}
	}
	\noindent where $\psi$ is the digamma function, $\psi_{T}$ is the multivariate digamma function and $\Gamma_{T}$ is the multivariate gamma function. Since the expectations of the sigmoid functions do not admit closed form solutions, we approximate the sigmoid functions by quadratic lower bounds:

	{\small 
		\begin{align}
			\label{eqn: quadratic approximation to sigmoid}
			& -\log\left( \exp\left( \innerprod{(\mb w_{t}\circ \mb z)}{\mb x_{t}^{i}} \right)+1 \right)  \nonumber \\ 
			&\geq -\log\left( \exp\left( \innerprod{(\mb w'_{t}\circ \mb z')}{\mb x_{t}^{i}} \right)+1 \right) \nonumber\\
			&\quad -\frac{\innerprod{\mb x_{t}^{i}}{\mb w_{t}\circ \mb z - \mb w'_{t} \circ \mb z'}}{\exp\left( -\innerprod{(\mb w'_{t}\circ \mb z')}{\mb x_{t}^{i}} \right)+1} \nonumber \\
			&\quad -\frac{1}{8} \left( \mb w_{t} \circ \mb z - \mb w'_{t}\circ \mb z' \right)^{\top} \mb x_{t}^{i} \left( \mb x_{t}^{i} \right)^{\top} \left( \mb w_{t}\circ \mb z - \mb w'_{t} \circ \mb z' \right).  
	\end{align}}
	\noindent for $i=1,\ldots, n_t, t=1, \ldots, T$, and $\mb w'_{t}, \mb z'$ are deterministic reference points of choice. This type of approximations has been used to design majorize-minimization (MM) algorithms for the logistic regression problem \cite{bohning1988monotonicity, hunter2004tutorial}. \rev{To clarify the tightness of this surrogate, let $f(\eta)=-\log(1+\exp(\eta))$. Since $f''(\eta)=-\sigma(\eta)\{1-\sigma(\eta)\}\geq -1/4$, the quadratic expression in Eqn.~\ref{eqn: quadratic approximation to sigmoid} is a global lower bound that is tangent at the reference value $\eta_0=\innerprod{(\mb w'_t\circ\mb z')}{\mb x_t^i}$. In the CAVI implementation, this reference point is updated using the variational mean parameters from the previous iteration, so the bound is repeatedly re-centered near the current region of the variational posterior. The resulting algorithm therefore optimizes a tractable surrogate lower bound rather than the exact ELBO; a nonzero gap can remain because the variational distribution has posterior variance and because later updates move away from the previous reference point.} The resulting approximation is, up to a constant:
	{
		\small 
		\begin{align}
			\label{eqn: elbo aprx}
			&-\sum_{t}\sum_{i}\expect[\mb W, \mb z\sim q]{\log\left( \exp\left( \innerprod{(\mb w_{t}\circ \mb z)}{\mb x_{t}^{i}} \right)+1 \right) } \nonumber \\
			&\geq -\sum_{t}\sum_{i} \log\left( \exp\left( \innerprod{(\mb w'_{t}\circ \mb z')}{\mb x_{t}^{i}}  \right)  +1 \right)  \nonumber \\
			&\quad +\sum_{t}\sum_{i} \frac{\innerprod{\mb w_{t}'\circ\mb z'}{\mb x_{t}^{i}}}{\exp\left(-\innerprod{(\mb w_{t}' \circ \mb z')}{\mb x_{t}^{i}}  \right)+1}\nonumber\\
			&\quad -\frac{1}{8} \sum_{t}\sum_{i} \innerprod{\mb w_{t}'\circ\mb z'}{\mb x_{t}^{i}}^{2}\nonumber \\
			&\quad -\sum_{t}\sum_{i} \frac{1}{\exp\left(-\innerprod{(\mb w'_{t}\circ \mb z')}{\mb x_{t}^{i}} \right)+1} \innerprod{\mb x_{t}^{i}}{(\mb m_{t} \circ \mb \phi)} \nonumber \\
			&\quad +\frac{1}{4} \sum_{t}\sum_{i} \innerprod{\mb w_{t}'\circ \mb z'}{\mb x_{t}^{i}} \innerprod{\mb m_{t} \circ \mb \phi}{\mb x_{t}^{i}}   \nonumber\\
			&\quad -\frac{1}{8}\sum_{t}\sum_{i} \innerprod{\mb m_{t} \circ \mb \phi}{\mb x_{t}^{i}}^{2}  \nonumber \\
			&\quad +\frac{1}{8} \sum_{t} \sum_{j} \left( m_{t,j}^2(\phi_{j}-1)-(\mb \Sigma_{j})_{t,t}  \right) \phi_{j} \sum_{i} (x_{t}^{ij})^{2}. 
		\end{align}
	}
	
	Alternative lower bounds are available \cite{jaakkola1997variational,drugowitsch2013variational,bishop2006pattern}, but they introduce additional variational parameters that scale with the feature dimension $d$. We avoid those formulations here in order to keep inference manageable for high-dimensional microbiome datasets.

	\subsubsection{Coordinate Ascent Variational Inference (CAVI)}
	\label{subsec: CAVI}
	
	Coordinate ascent variational inference (CAVI)  \cite{bishop2006pattern,blei2017variational} optimizes one block of variational parameters at a time while holding the remaining blocks fixed. In this setting, that strategy yields an iterative algorithm that alternates among updates for the coefficient means and covariances, the feature-inclusion probabilities, and the global hyperparameters. The resulting procedure is substantially more scalable than sampling-based alternatives for the high-dimensional microbiome setting considered here.
	
	One useful result (Equation 18 of \cite{blei2017variational}) states that if we are to approximate a general posterior distribution $p\left(  \mb \xi \mid \text{data}  \right) $ with a mean-field approximation $q\left(  \mb \xi \right):=\prod_{j} q_{j}\left(\xi_{j} \right) $, the CAVI update for $j$-th latent variable $\xi_{j}$ (i.e., the optimal solution $q_{j}^{\star}\left( \xi_{j} \right) $) is proportional to the exponentiated conditional expected log of the joint:
	\begin{equation}
		\label{eqn: cavi update rule}
		q^{\star}_{j} \propto \exp\left( \expect[\mb \xi_{-j}\sim q_{-j}]{\log\left( p\left( \xi_{j}, \mb \xi_{-j}  \mid  \text{data}  \right)  \right) } \right).	
	\end{equation}
	where $\mb \xi_{-j}$ corresponds to all but the $j$-th latent variable.  
	
	A standard result in variational inference states that, under a mean-field factorization, each variational factor is updated by exponentiating the conditional expectation of the joint log-density with respect to the remaining factors. We use this identity to derive the updates summarized in Algorithm \ref{alg: CAVI update summary}; full derivations are provided in Section 2 of the Supplementary Information document.
	
	\begin{algorithm}[ht]
		\footnotesize
		\caption{CAVI for Bayesian Multitask Sparse Logistic Regression}
		\label{alg: CAVI update summary}
		\begin{algorithmic}[1]
			\Procedure {CAVI}{Inputs: $\left(\mb x_{t}^{i}, y_{t}^{i}\right)_{t=1,\ldots T, i=1,\ldots n_{t}}$}
			\ForAll {$\text{itr}=1,\ldots, \text{Niter}$}
			\ForAll{$j=1,\ldots d$} 
			\State $\tilde{\mb X}_j \leftarrow \diag\left(\left[
			\begin{array}{c}
				\sum_{i} \phi_{j}(x_{1}^{ij})^2\\
				\vdots \\
				\sum_{i} \phi_{j}(x_{T}^{ij})^2
			\end{array}
			\right]\right)$
			\Statex
			\Statex
			\State $\mb \Sigma_{j} \leftarrow \left( v \mb V + \frac{1}{4} \tilde{\mb X_{j}} \right)^{-1}$
			\EndFor
			\ForAll{$j=1,\ldots,d$} 
			\ForAll{$t=1,\ldots,T, i=1,\ldots, n_t$} 
			\State $\tilde{y_t^i} \leftarrow \text{sigmoid}\left(\sum_j m_{tj} x_t^{ij}\phi_j\right)$
			\EndFor
			\State $\mb m_{\left( j \right) } \leftarrow \mb \Sigma_{j} \left[
			\begin{array}{c}
				\phi_{j}\sum_{i} \left(y_{1}^{i}-\tilde{y_{1}^{i}}\right) x_1^{ij} \\
				\vdots \\ 
				\phi_{j}\sum_{i} \left(y_{T}^{i}-\tilde{y_{T}^{i}}\right) x_{T}^{ij} 
			\end{array}\right]$ 
			\Statex \hspace{\algorithmicindent}$\phantom{\mb m_{\left( j \right) } \leftarrow } \qquad +\frac{1}{4} \mb \Sigma_{j} \left[
			\begin{array}{cc}
				\phi_{j}^{2} \sum_{i} \left(x_{1}^{ij}\right)^{2} m_{1j}  \\
				\vdots \\
				\phi_{j}^{2} \sum_{i} \left( x_{T}^{ij} \right)^{2} m_{Tj}
			\end{array}\right]$
			\EndFor
			\ForAll{$j=1,\ldots, d$} 
			\ForAll{$t=1,\ldots,T, i=1,\ldots, n_t$} 
			\State $\tilde{y_t^i} \leftarrow \text{sigmoid}\left(\sum_j m_{tj} x_t^{ij}\phi_j\right)$
			\EndFor
			\State $\phi_j \leftarrow \text{sigmoid} \Bigl( \psi\left(\alpha\right) - \psi\left(\beta\right) $
			\Statex \hspace{\algorithmicindent}$\phantom{\phi_j \leftarrow \text{sigmoid}(} + \sum_{t}\sum_{i} (y_{t}^{i} - \tilde{y}_{t}^{i})m_{tj} x_{t}^{ij}$
			\Statex \hspace{\algorithmicindent}$\phantom{\phi_j \leftarrow \text{sigmoid}(} + \frac{1}{4}\sum_{t}(m_{tj}^{2} \phi_j)\sum_{i}(x_{t}^{ij})^2$
			\Statex \hspace{\algorithmicindent}$\phantom{\phi_j \leftarrow \text{sigmoid}(} - \frac{1}{8}\sum_{t}m_{tj}^{2} \sum_{i}(x_{t}^{ij})^2$
			\Statex \hspace{\algorithmicindent} $\phantom{\phi_j \leftarrow \text{sigmoid}(} - \frac{1}{8}\sum_{t}(\Sigma_j)_{tt} \sum_{i}(x_{t}^{ij})^2 \Bigr)$
			\EndFor
			\State $\alpha \leftarrow \alpha_0+\sum_{j} \phi_{j}$, 
			\State $\beta \leftarrow \beta_0+d-\sum_{j} \phi_{j}$, 
			\State $\theta \leftarrow \frac{\alpha}{\alpha+\beta}$, 
			\State $v \leftarrow  v_0+d$,
			\State $\mb V \leftarrow \left( \mb V_0^{-1}+\sum_{j} \mb m_{\left( j \right)} \mb m_{\left( j \right) }^{\top} + \mb \Sigma_{j} \right)^{-1}$
			\EndFor
			\EndProcedure
		\end{algorithmic}
	\end{algorithm}

	\rev{The CAVI iterations are stopped when the relative change in the variational surrogate objective satisfies}
	\begin{equation}
		\frac{|\mathcal L_{\mathrm{surr}}^{(k)}-\mathcal L_{\mathrm{surr}}^{(k-1)}|}{1+|\mathcal L_{\mathrm{surr}}^{(k-1)}|}\leq 10^{-4},
	\end{equation}
	\rev{with a maximum iteration cap used as a safeguard. In the released simulated-data workflow, the final BayesMTL fit uses a cap of 100 CAVI iterations, while stochastic hyperparameter-search fits use 40 iterations and at most 4000 sampled feature updates per stochastic step. With cached linear predictors or residuals, a full CAVI iteration for one task cluster with $d$ features, $T$ tasks, and total sample size $N=\sum_t n_t$ has per-iteration cost $O(dN+dT^3)$; the $O(dT^3)$ term comes from the feature-wise inversions of $T\times T$ covariance matrices. A literal implementation that recomputed each linear predictor from scratch inside every feature update would add an $O(d^2N)$ cost, which is avoided by cached predictor/residual updates. For multiple disease clusters, the corresponding cost is $\sum_c O(d_cN_c+d_cT_c^3)$.}
	
	\subsection{Experimental setup}
	\label{subsec: experimental setup}
	
	We evaluate the proposed method on both synthetic and real microbiome datasets. The synthetic experiments are designed to test whether the model can recover shared sparse structure under controlled conditions, whereas the real-data experiments assess predictive behavior, calibration, and feature interpretability on pooled studies. We compare BayesMTL with single-task, pooled, and multitask baselines that represent common alternatives for sparse classification, which are described below.
	
	\begin{enumerate}
		\item Single-task Logistic Classifier with sparsity penalty (STL-LC) \cite{tibshirani1996regression,banerjee2008model}: this is a single-task model, where we fit independent logistic models to each task separately. This is an extension of standard LASSO to the binary classification setting. 
		\item Pooled Logistic Classifier with sparsity penalty (Pooled-LC): we train a single logistic regression model on the combined data from all datasets. 
		\item MTFL (Multitask Feature Learning) \cite{liu2009multi}: this is an optimization based approach to multitask learning based on $\ell_{2,1}$-norm regularization. The proposed method can be seen as a Bayesian Hierarchical extension, and the difference in modeling is manifested in how the sparsity pattern is encouraged: we use an overparameterized Bernoulli-Gaussian model, which has better support recovery properties \cite{bazot2011bernoulli}.
		\item MSSL (Multitask Sparse Structure Learning) \cite{gonccalves2016multi}: this is an optimization based multitask learning approach, where the imposed sparsity structure is on the precision (inverse covariance matrix) of the regression coefficients across tasks. The optimization problem of this formulation is equivalent to the graphical lasso problem \cite{friedman2008sparse,mazumder2012graphical} for covariance estimation. 
	\end{enumerate}
	
	\rev{Hyperparameters for all methods are selected on the training data by repeated stratified validation. Each trial uses 20 repeated stratified 80/20 validation splits, preserving class balance within each task, and the objective is validation log predictive likelihood, equivalently minimization of validation cross-entropy. The same split-generation protocol and random seed are used across methods. Hyperparameter optimization uses 200 Optuna/Ray Tune trials. For BayesMTL, $\alpha_0\in[1,10]$, $\beta_0\in[1,100]$, and $v_0\in[1,10]$ are sampled on a log-uniform scale, with the weighted construction of $\mb V_0$ used in the reported experiments. For Pooled-LC, the $\ell_1$ penalty weight is searched over $[10^{-6},10^3]$ on a log-uniform scale. For STL-LC, independent task-specific logistic models use an $\ell_1$ penalty with $C\in[10^{-3},1]$ sampled uniformly. For MTFL, $\rho_{2,1}$ and $\rho_2$ are each searched over $[0.1,3000]$ on a log-uniform scale. For MSSL, $\lambda_1\in[0.1,1000]$ and $\lambda_2\in[10^{-5},100]$ are searched on log-uniform scales. For multitask methods, if validation scores are within a 1\% relative tolerance of the best score, the sparser model is selected as a secondary preference. We report balanced accuracy as the primary predictive metric because it is more informative than raw accuracy under class imbalance. Additional implementation details are provided in the Supplementary Information section titled ``Experimental setup and reproducibility details''.} \revii{Hyperparameter selection uses only validation splits drawn from the training partition, disjoint from the held-out test data, so the reported test-set metrics are not affected by selection-induced optimism.}
	
	\section{Results and Discussion}
	\label{sec: bmtl Experiments}

	\subsection{Synthetic Datasets}
	\label{subsec: synthetic datasets}
	
	We generate synthetic datasets to examine whether our algorithm is able to: 1) recover the support of the regression coefficients that correspond to informative features for the prediction, which are evaluated by balanced accuracy; 
	2) recover the ground truth regression coefficients (up to normalization), which are evaluated by cosine distance (i.e., one minus the normalized inner product). 
	
	The synthetic experiments are designed to isolate the conditions under which the proposed multitask prior should help. Specifically, they test whether BayesMTL can recover the correct sparse support and the underlying coefficient structure when tasks truly share informative features. By varying both sparsity level and sample-size imbalance, these simulations probe performance across easier and harder multitask settings. \rev{Each simulation uses $T=10$ tasks, $d=100$ features, and 10 independent runs. Conditional on a shared support vector $z_j^*\stackrel{\mathrm{i.i.d.}}{\sim}\mathrm{Bernoulli}(\theta)$, task-specific coefficients are drawn from a shared covariance model and labels are sampled from $y_t^i\sim\mathrm{Bernoulli}[\mathrm{sigmoid}\{(\mb x_t^i)^\top(\mb w_t\circ\mb z^*)\}]$. Training and test covariates are generated independently from $\mathcal N_d(\mb 0,\mb I_d)$, conditional on the same simulation parameters.} Additional details of the synthetic datasets designed for this experiment is presented in Table \ref{tab: simulation setup} and the Supplementary Information section titled ``Experimental setup and reproducibility details''. \rev{The simulation design varies sparsity and unequal task information rather than exhaustively varying all possible heterogeneity factors. In a first-order logistic-regression approximation, coefficient-recovery difficulty is governed by terms such as $n_t\beta_{tj}^2$, so the negative-binomial sample-size regime probes an effective signal-to-noise and estimation-noise setting through unequal sample sizes across tasks.} \revii{These synthetic data are generated under the model's assumptions by design, so as to validate support and coefficient recovery under correct specification; robustness under misspecification is instead probed by the real-data analysis, which is not model-generated. Benchmarking against external microbiome-specific simulators is left as future work.} 
	
\begin{figure*}[htbp]
	\centering
	
	\begin{subfigure}{0.45\textwidth}
		\centering
		\includegraphics[width=\textwidth]{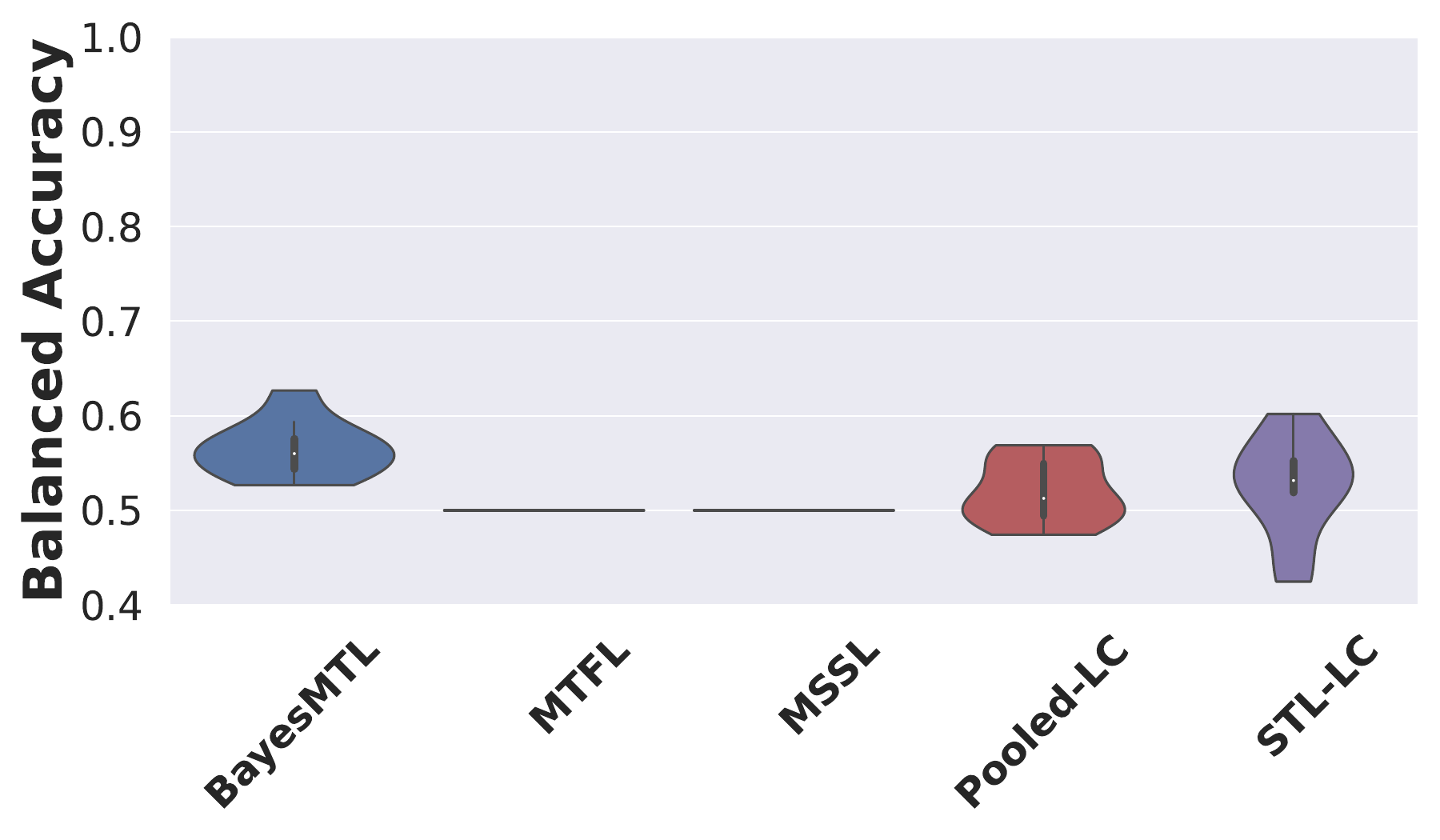}
		\caption{Dataset1: balanced and dense}
	\end{subfigure} \hfill
	\begin{subfigure}{0.45\textwidth}
		\centering
		\includegraphics[width=\textwidth]{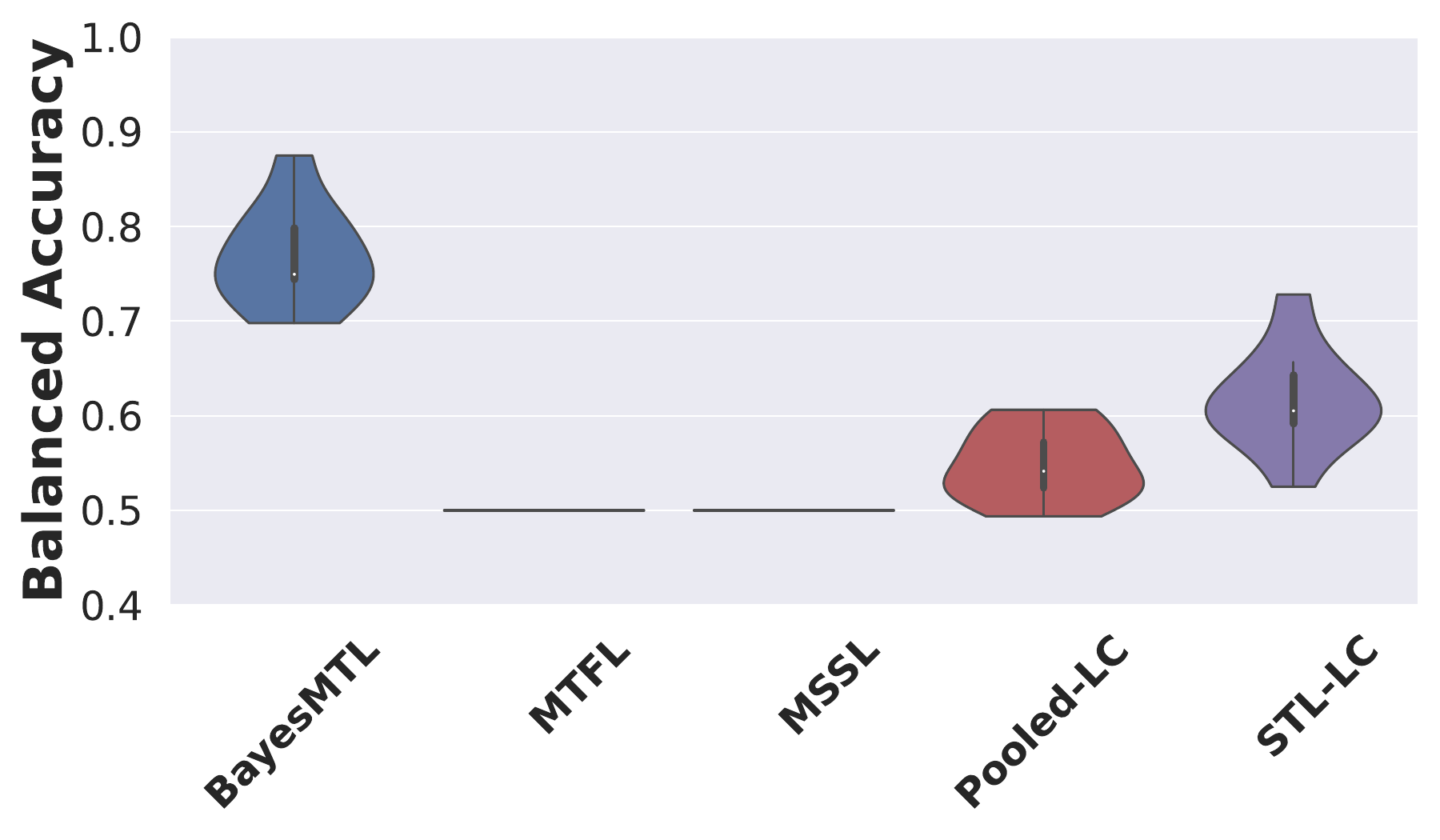}
		\caption{Dataset2: balanced and sparse}
	\end{subfigure}
	
	\begin{subfigure}{0.45\textwidth}
		\centering
		\includegraphics[width=\textwidth]{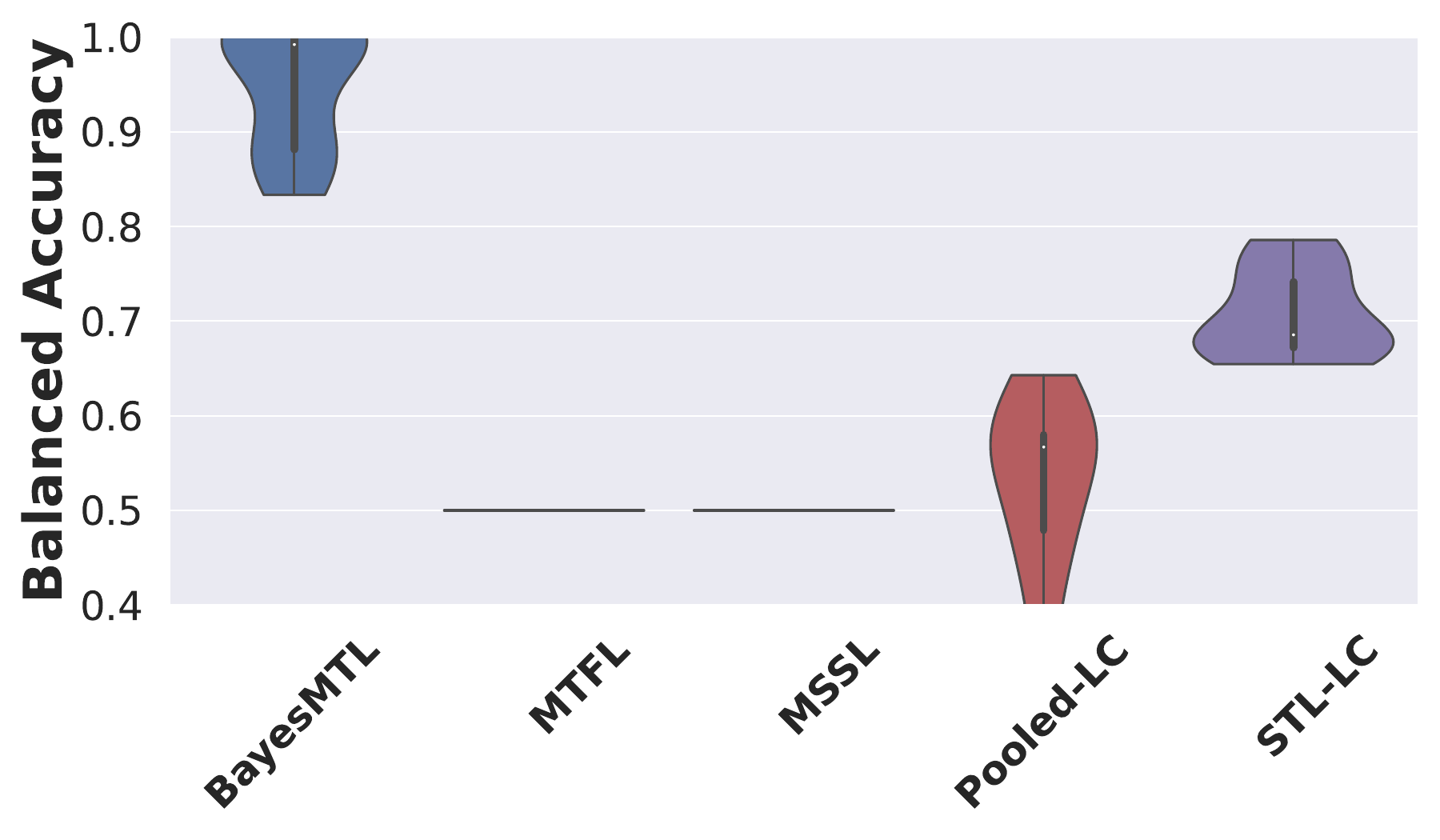}
		\caption{Dataset3: balanced and ultra sparse}
	\end{subfigure} \hfill
	\begin{subfigure}{0.45\textwidth}
		\centering
		\includegraphics[width=\textwidth]{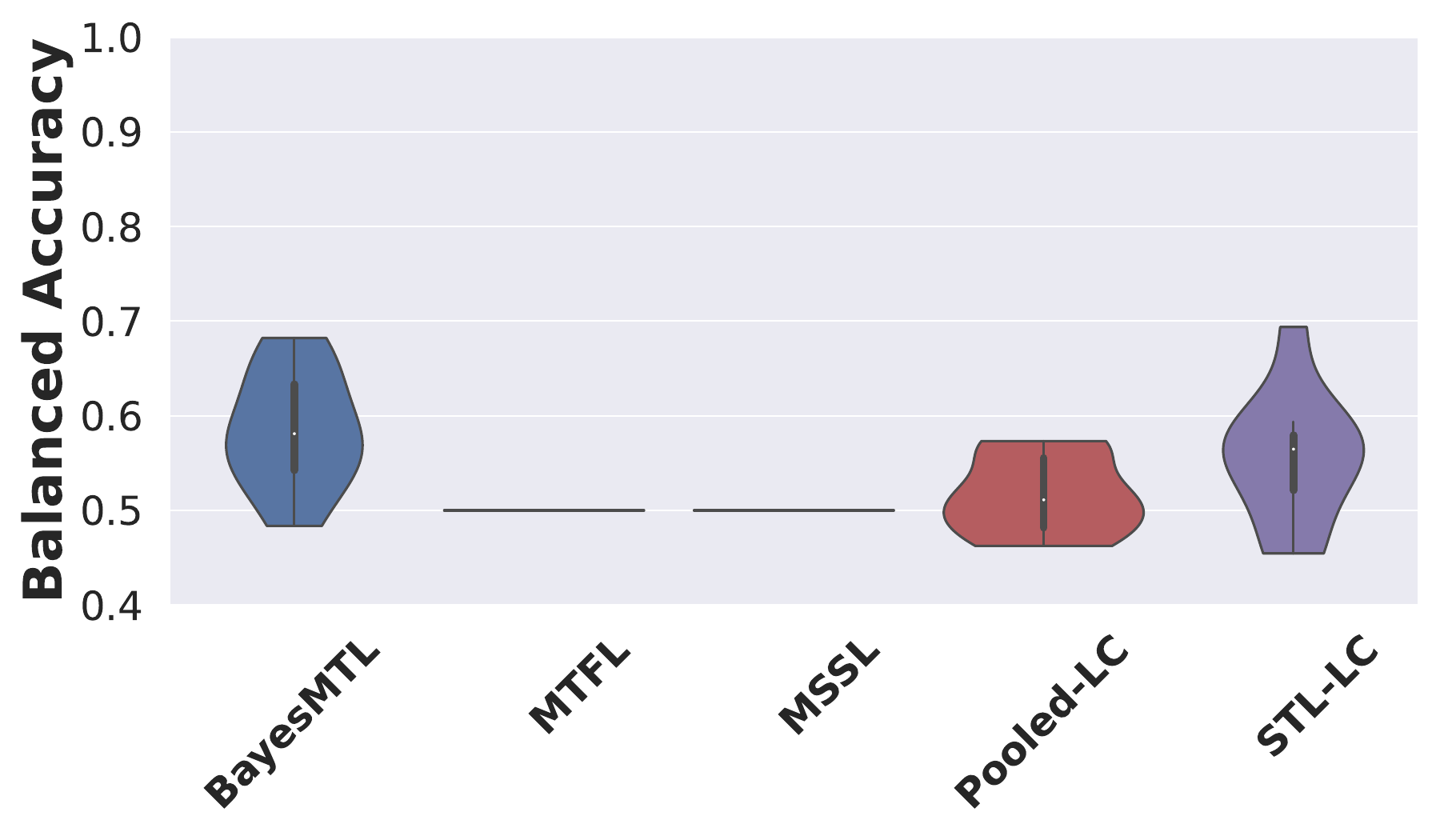}
		\caption{Dataset4: imbalanced and dense}
	\end{subfigure}
	
	\begin{subfigure}{0.45\textwidth}
		\centering
		\includegraphics[width=\textwidth]{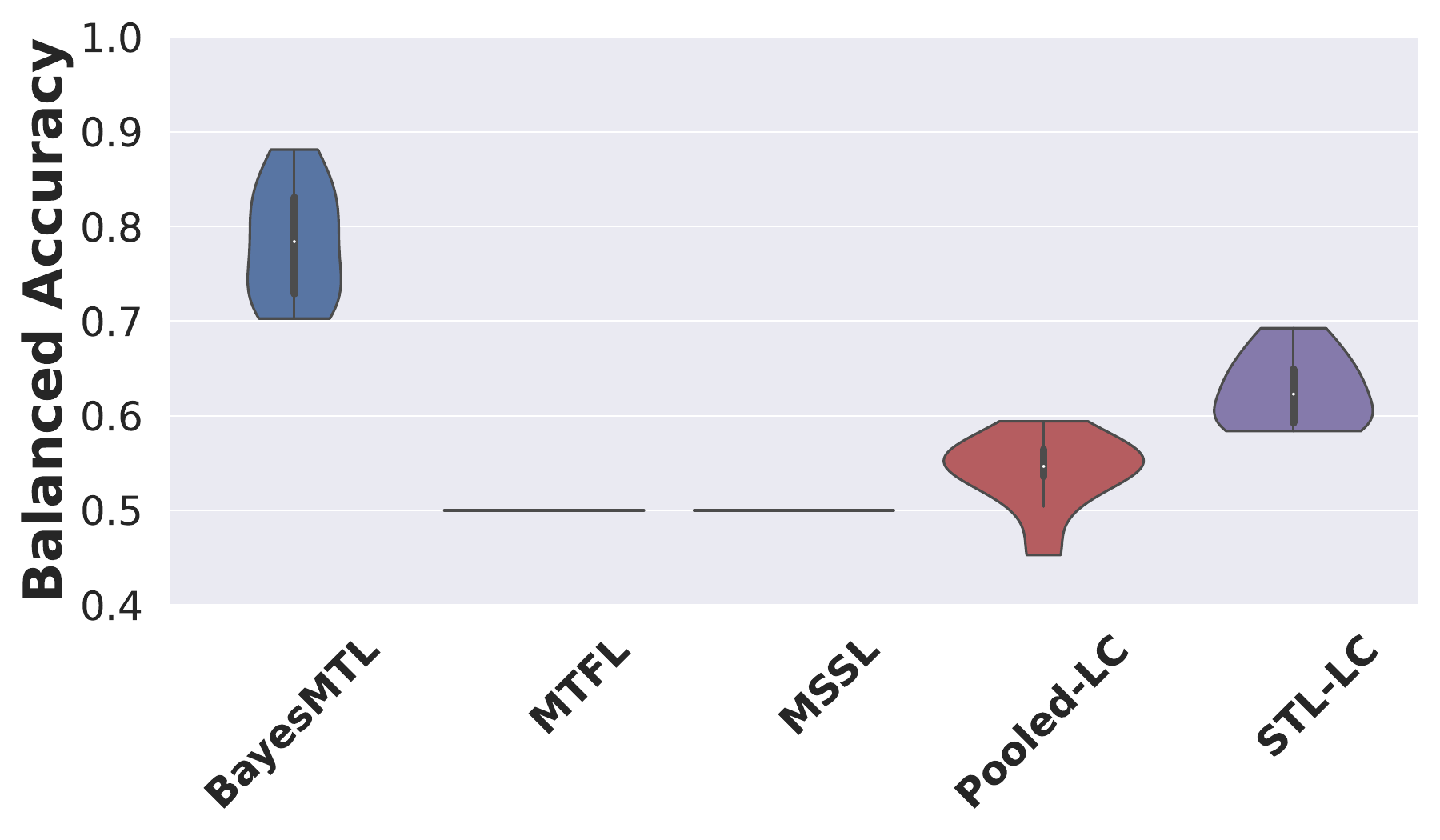}
		\caption{Dataset5: imbalanced and sparse}
	\end{subfigure} \hfill
	\begin{subfigure}{0.45\textwidth}
		\centering
		\includegraphics[width=\textwidth]{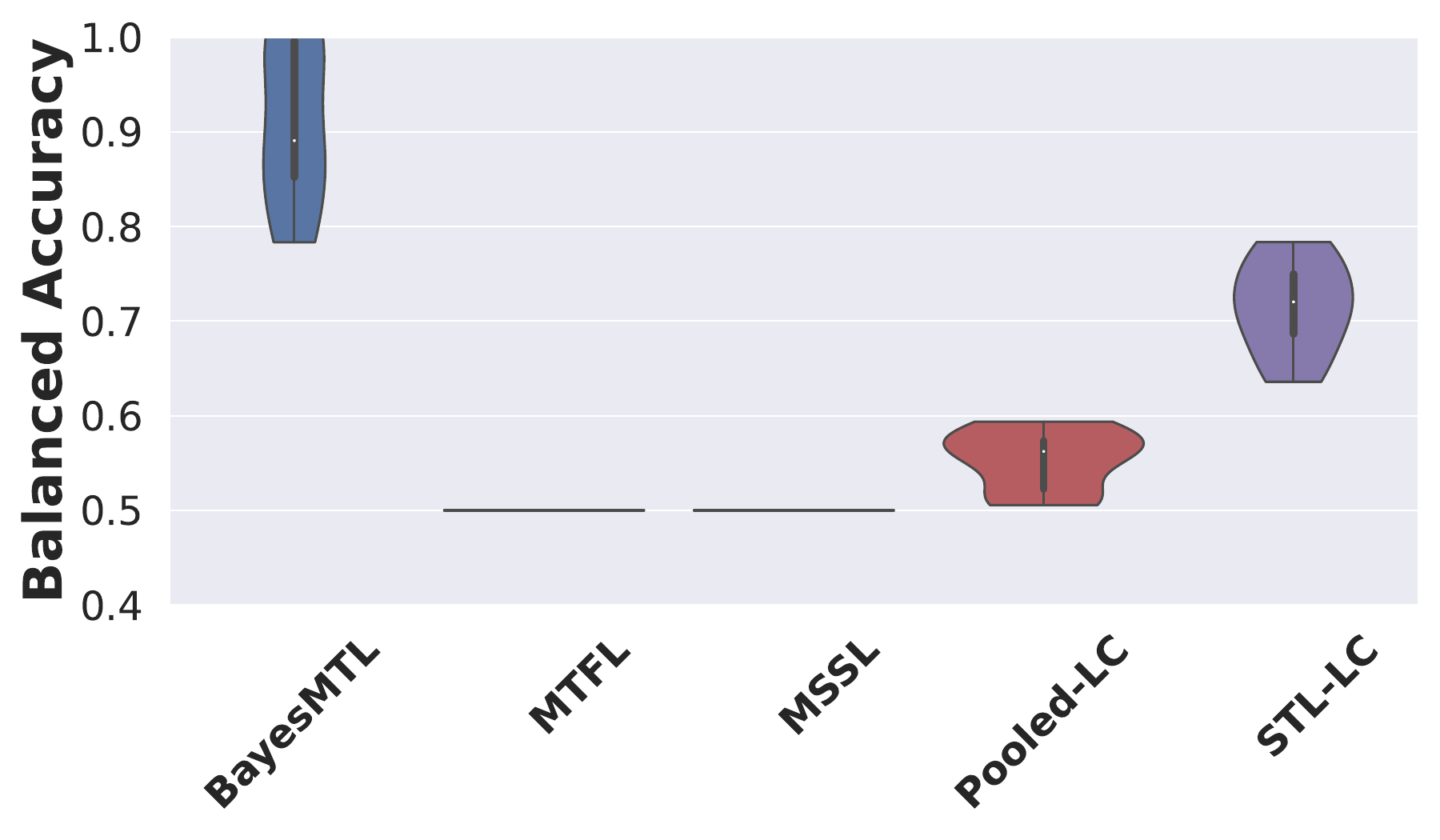}
		\caption{Dataset6: imbalanced and ultra sparse}
	\end{subfigure}
	
	\caption{Summary of the support recovery results for the simulated data evaluated using balanced accuracy across \rev{10 independent runs}. BayesMTL consistently outperforms baseline methods, particularly when shared sparsity exists across regression coefficients.}
	\label{fig:simulated_support_recovery}
\end{figure*}

	\textbf{Support recovery:} We convert support recovery into a binary prediction problem in which each feature is classified as either relevant or irrelevant. Figure \ref{fig:simulated_support_recovery} shows that the proposed Bayesian method performs best in most settings, with particularly strong performance when the true coefficients share a sparse support.
	
	\textbf{Weight recovery:} We also assess how accurately each method recovers the underlying regression coefficients. Because logistic predictions are invariant to global scaling, we use cosine distance as the evaluation metric. Table \ref{tab: simulated weight recovery} shows that the proposed Bayesian method performs best in all but the dense setting.

	\begin{table}[htbp]
		\centering
		\begin{tabular}{ccc}\toprule
			& {Balanced} & {Imbalanced}\\ \midrule
			{Dense} $\left(\theta=0.8\right)$ & {Dataset-1} & {Dataset-4} \\ 
			{Sparse} $\left(\theta=0.2\right)$ & {Dataset-2} & {Dataset-5} \\
			{Ultra Sparse} $\left( \theta=0.05 \right)$ & {Dataset-3} & {Dataset-6} \\  \bottomrule
		\end{tabular}
		\caption{\rev{Summary of the simulated datasets. Balanced sample sizes are generated as $n_t=16+\mathrm{Poisson}(24)$, and imbalanced sample sizes are generated as $n_t=16+\mathrm{NegBin}(r=1,p=0.04)$, where the negative-binomial variable counts failures before the first success. Thus both settings have expected per-task sample size $40$, while the negative-binomial setting has substantially larger across-task variability. For all simulations, the number of features is $d=100$, the number of tasks is $T=10$, and $\theta$ is the probability that a feature belongs to the true predictive support.}}
	\label{tab: simulation setup}
\end{table}

\begin{table}[htbp]
	\centering
	\resizebox{0.99\columnwidth}{!}{
		\begin{tabular}{l ccccc}
			\toprule
			& {BayesMTL} & {MTFL} & {MSSL} & {Pooled-LC} & {STL-LC} \\ 
			\midrule  	
			Dataset-1 & 0.74 (0.06) & 0.56 (0.02) & \bfseries 0.55 (0.02) & 0.98 (0.04) & 0.73 (0.02) \\
			Dataset-2 & \bfseries 0.34 (0.09) & 0.56 (0.01) & 0.56 (0.01) & 1.02 (0.05) & 0.56 (0.05) \\ 
			Dataset-3 & \bfseries 0.10 (0.04) & 0.59 (0.03) & 0.59 (0.02) & 0.99 (0.12) & 0.26 (0.04) \\ 
			Dataset-4 & 0.67 (0.10) & 0.58 (0.08) & \bfseries 0.58 (0.05) & 1.02 (0.03) & 0.73 (0.07) \\ 
			Dataset-5 & \bfseries 0.46 (0.11) & 0.59 (0.08) & 0.57 (0.04) & 0.57 (0.09) & 1.04 (0.06) \\ 
			Dataset-6 & \bfseries 0.22 (0.12) & 0.65 (0.05) & 0.61 (0.04) & 1.07 (0.10) & 0.35 (0.11) \\ 
			\bottomrule
	\end{tabular}}
	\caption{Summary of the weights recovery measured in cosine distance. The cosine distance is bounded between $0$ and $2$ with $0$ meaning perfect recovery. \rev{Values in parentheses are standard deviations over 10 independent runs.} The proposed Bayesian (BayesMTL) approach outperforms the benchmark methods when there is a shared sparsity structure across regression coefficients of different tasks.}
	\label{tab: simulated weight recovery}
\end{table}

\rev{For the simulation summaries, a standard error can be obtained from a reported standard deviation $s$ as $\mathrm{SE}=s/\sqrt{10}$. A 95\% confidence interval for a reported mean can be computed as $\bar{x}\pm t_{0.975,9}s/\sqrt{10}$.} \revii{These simulations are intended to demonstrate support recovery when a shared sparse structure is present rather than to provide precise performance estimates. In the sparse and ultra-sparse settings, where such a structure exists, the recovery advantage over the baselines is large relative to the run-to-run variability, so ten runs is sufficient to establish this conclusion.}

\subsection{Microbiome Data}
\label{subsec: microbiome data}

In this subsection, we apply the proposed model to previously published microbiome datasets described in Section \ref{sec: intro}. We aim to assess whether leveraging shared structure across multiple diseases improves classification performance, and whether the model can highlight microbial taxa associated with each disease.

\begin{figure}[htbp]
	\centering
	\begin{subfigure}{0.45\textwidth}
		\centering		
		\includegraphics[width=\textwidth]{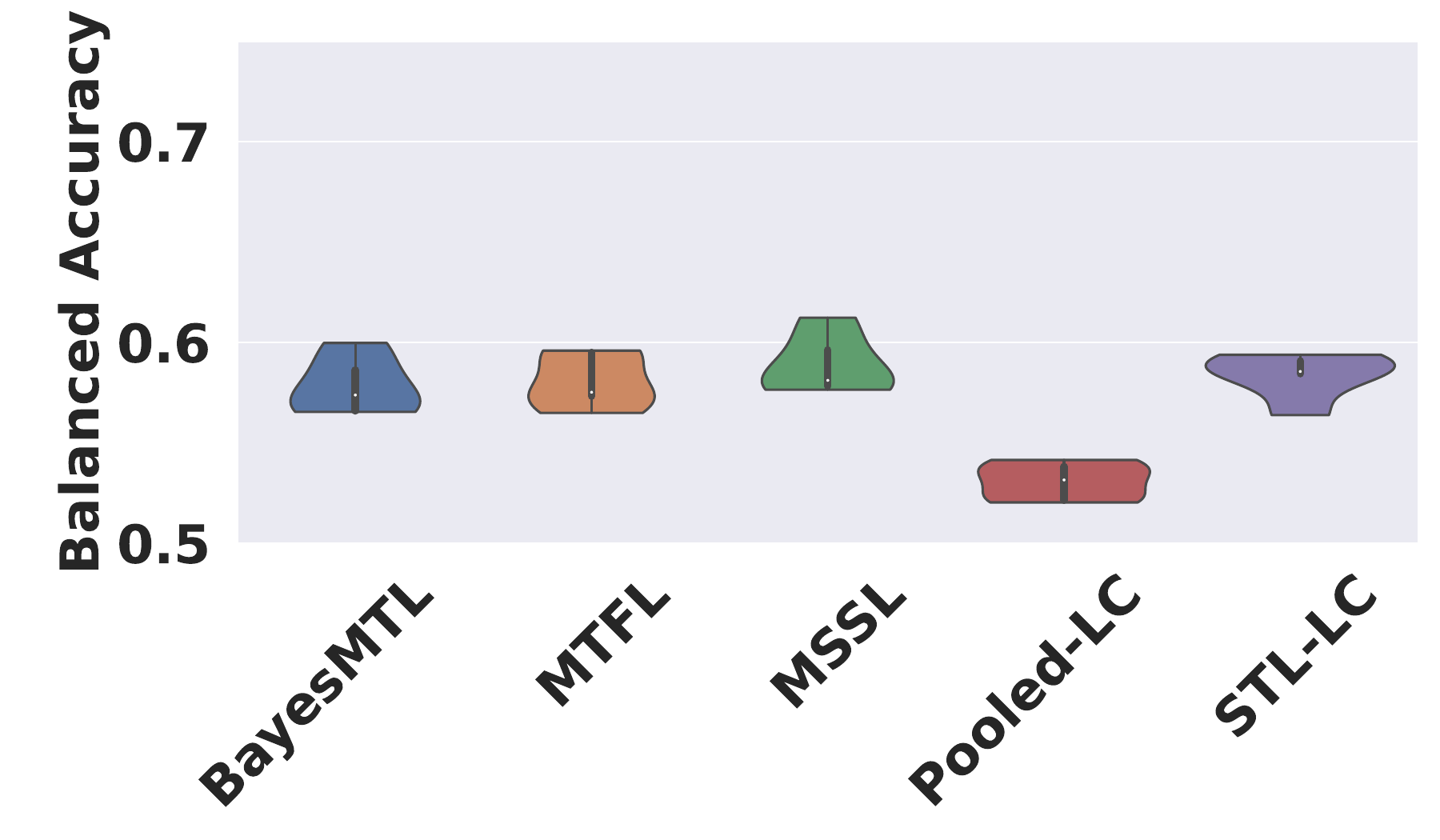}
		\caption{\centering Kingdom} 
	\end{subfigure}
	\begin{subfigure}{0.45\textwidth}
		\centering
		\includegraphics[width=\textwidth]{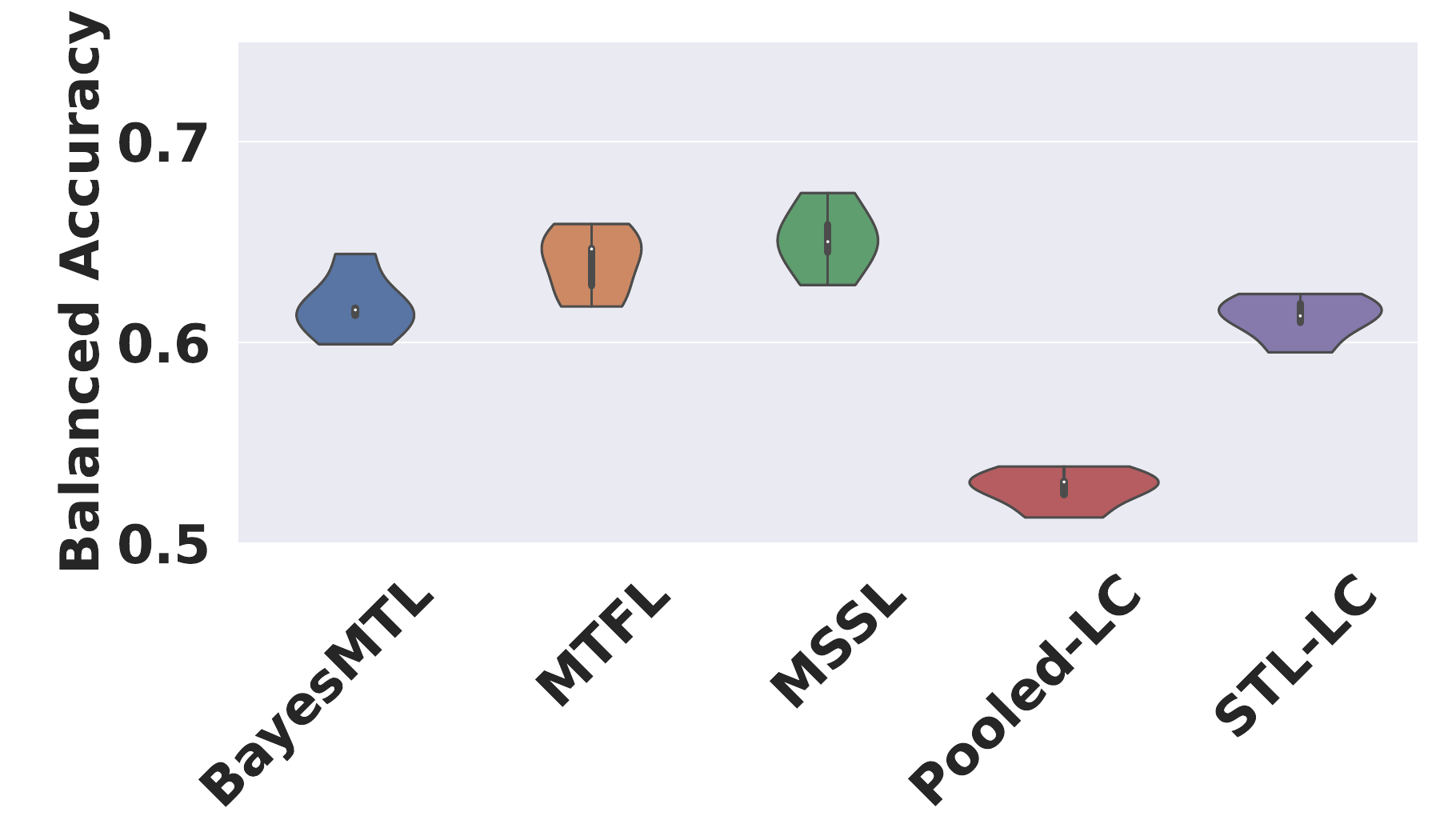}
		\caption{\centering Phylum}
	\end{subfigure}   
	\begin{subfigure}{0.45\textwidth}
		\centering
		\includegraphics[width=\textwidth]{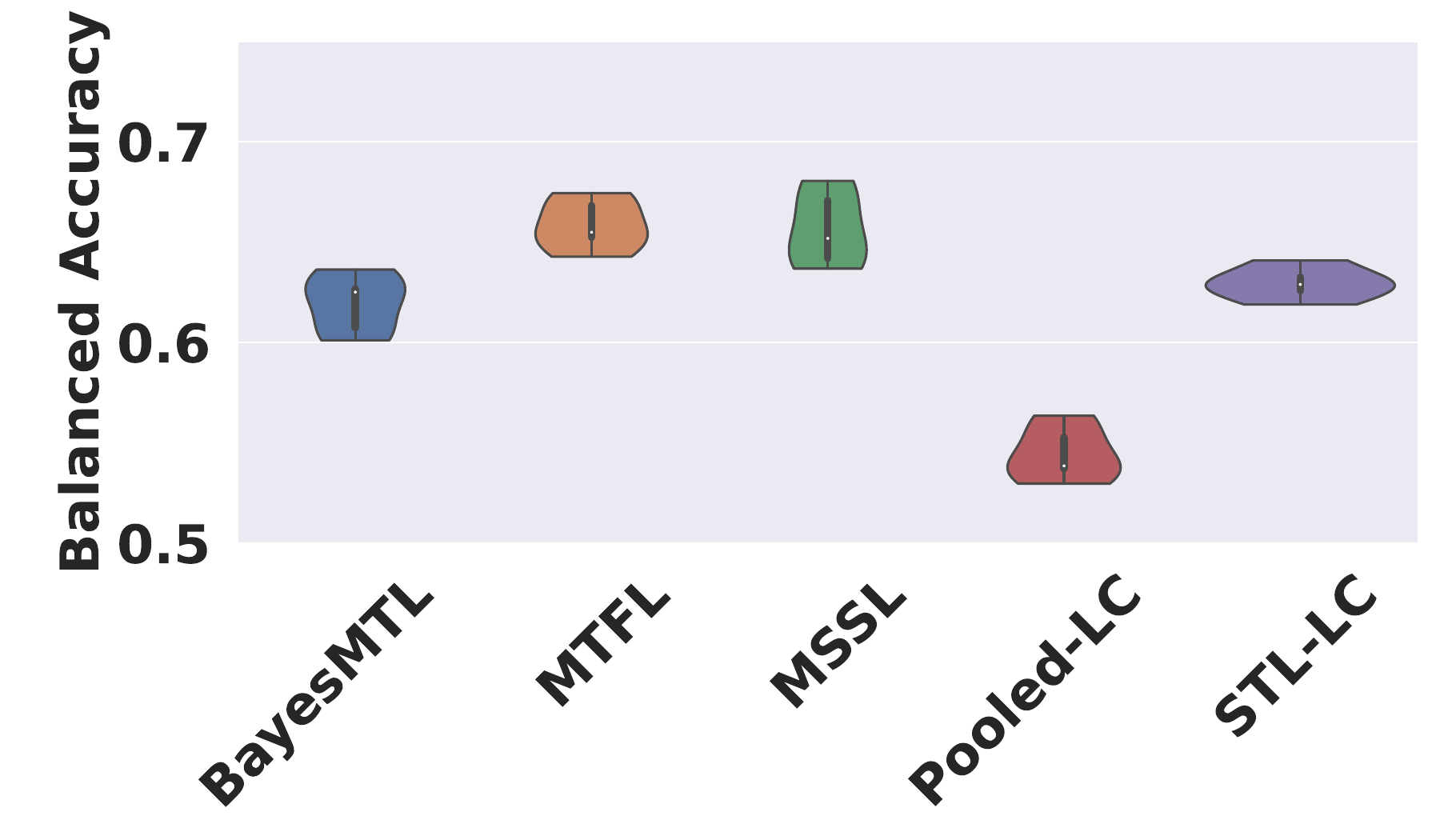}
		\caption{\centering Class}
	\end{subfigure}
	\begin{subfigure}{0.45\textwidth}
		\centering		\includegraphics[width=\textwidth]{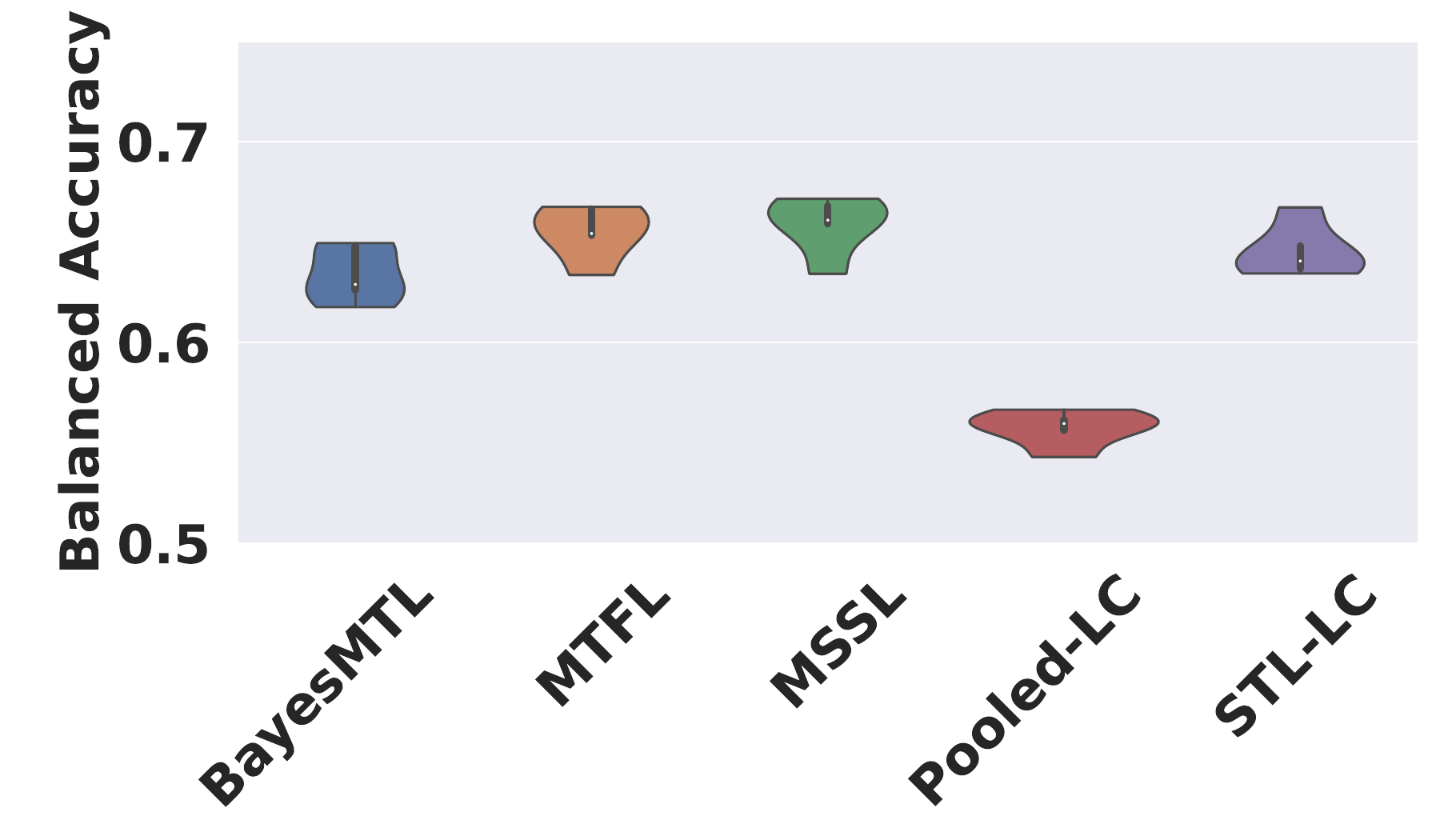}
		\caption{\centering Order}
	\end{subfigure}
	
	\begin{subfigure}{0.45\textwidth}
		\centering
		\includegraphics[width=\textwidth]{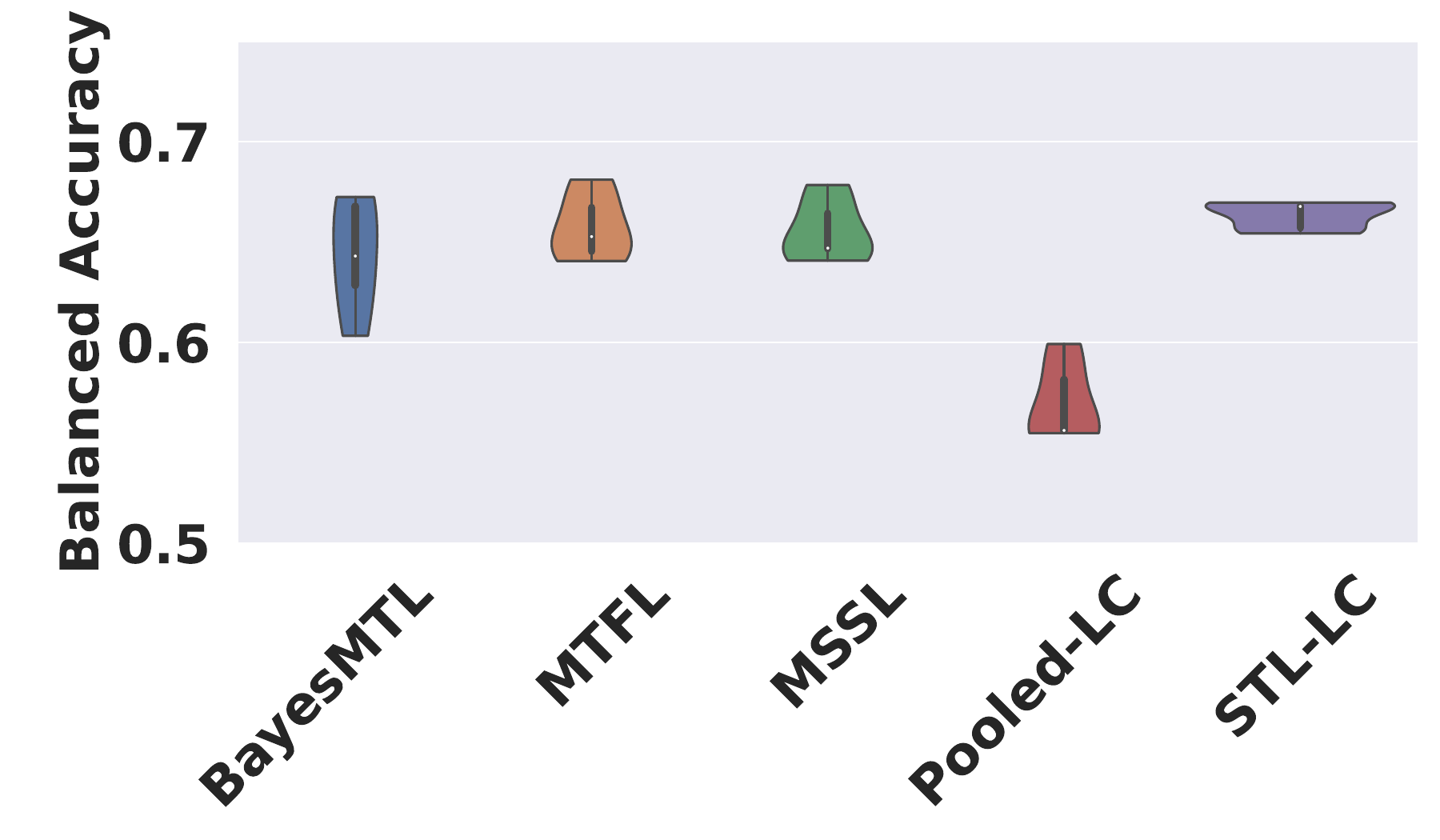}
		\caption{\centering Family}
	\end{subfigure}
	\begin{subfigure}{0.45\textwidth}
		\centering		\includegraphics[width=\textwidth]{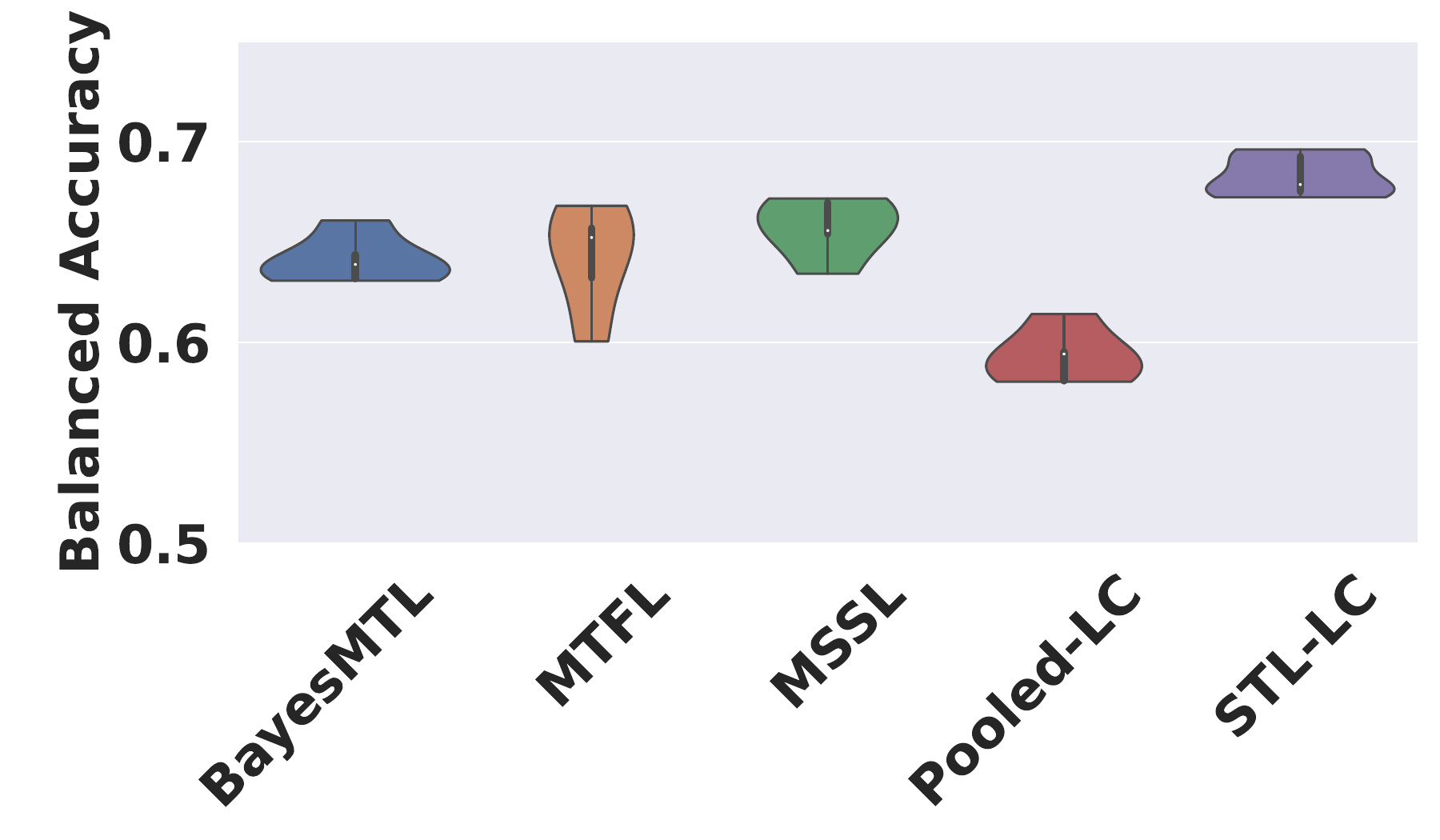}
		\caption{\centering Genus}
	\end{subfigure}     
	\begin{subfigure}{0.45\textwidth}
		\centering
		\includegraphics[width=\textwidth]{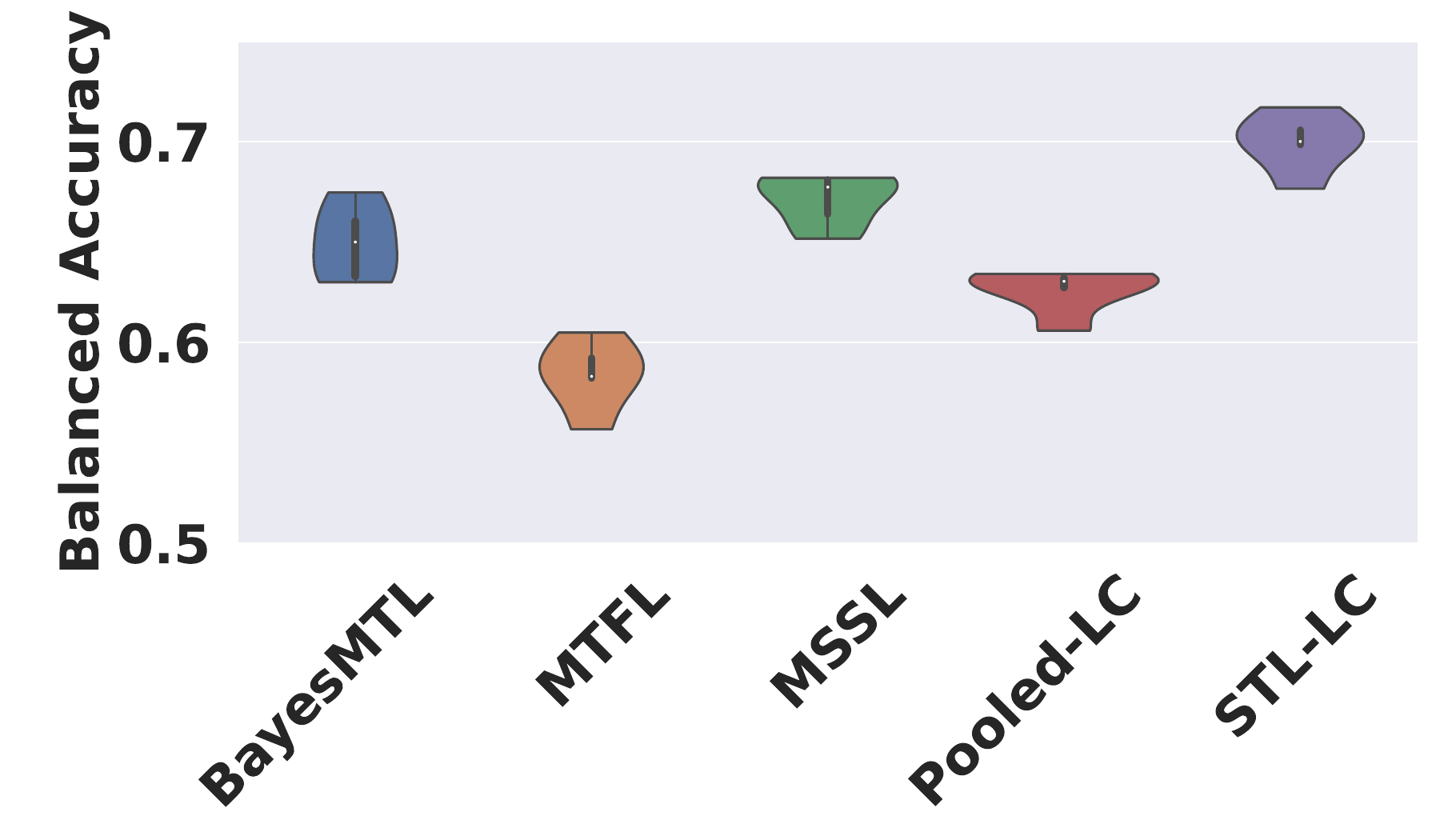}
		\caption{\centering Species}
	\end{subfigure}
	\caption{\rev{Summary of prediction performance evaluated by balanced accuracy across 7 taxonomic ranks. Due to the heterogeneous nature of the data, BayesMTL is competitive but not uniformly superior to independent single-task models. Its main advantage in this experiment is sparse multitask posterior inference: it identifies shared disease-category signals while retaining task-specific effects and uncertainty summaries.}}
	\label{fig: microbiome prediction}
\end{figure}

Figure \ref{fig: microbiome prediction} summarizes balanced accuracy across taxonomic ranks. We observe that jointly modeling diseases tends to be more beneficial at coarser taxonomic levels, where shared patterns across studies are more likely to emerge, although these improvements are not uniform across all datasets. This variability is consistent with the substantial heterogeneity in study design, patient populations, and sequencing protocols.
\rev{These real-data results should therefore not be interpreted as uniform predictive superiority over STL-LC. Independent single-task sparse logistic models can perform well when disease signals are study-specific. The main advantage of BayesMTL in this setting is the joint posterior over a shared cross-study inclusion structure, task-specific coefficients, and uncertainty-aware feature interpretation.} \revii{Concretely, the practical value of the multitask structure in this application is data-driven biological feature identification and hypothesis generation rather than improved classification. Because BayesMTL shares a single inclusion structure across studies of the same disease, it yields a cross-study-consistent set of candidate taxa; independent single-task models cannot express such a shared signature by construction, since each single-task fit selects features from one study in isolation. The diarrhea case study in Section~\ref{subsubsec: diarrhea biology} illustrates the biological coherence of the resulting shared signatures.}
Beyond predictive performance, a key advantage of the proposed model is its ability to produce sparse and interpretable solutions. \rev{As shown in Table \ref{tab: microbiome sparsity ratios}, BayesMTL selects substantially fewer features than the dense multitask baselines MTFL and MSSL while maintaining competitive accuracy, although STL-LC can be sparser at several ranks because it fits each study independently.} In the context of microbiome analysis, this is particularly valuable, as it narrows the set of candidate taxa for downstream biological investigation.

\begin{table}[htbp]
	\centering
	\resizebox{\columnwidth}{!}{
		\begin{tabular}{lccccc}\toprule
			Taxon Ranks &    BayesMTL &          MTFL &            MSSL &       Pooled-LC &          STL-LC \\ \midrule
			Kingdom  &  0.29 (0.01) & 0.54 (0.03) &  0.47 (0.03) &  0.88 (0.04) &          \bf{0.26} (0.02) \\
			Phylum &            0.18 (0.02) & 0.30 (0.002) &  0.31 (0.003) &  0.78 (0.03) & \bf{0.086} (0.02) \\
			Class &             0.12 (0.01) & 0.24 (0.006) &  0.30 (0.05) &  0.69 (0.02) & \textbf{0.070} (0.005) \\
			Order &             0.11 (0.01) & 0.20 (0.01) &  0.22 (0.01) &  0.54 (0.009) & \textbf{0.046} (0.008)  \\
			Family &    0.043 (0.02) & 0.41 (0.3) &  0.17 (0.002) &  0.41 (0.003) &         \bf{0.041} (0.007) \\
			Genus &     \bf{0.018} (0.007) & 0.40 (0.3) &  0.15 (0.0001) &  0.33 (0.003) &  0.035 (0.006) \\
			Species &   \bf{0.029} (0.005) & 0.15 (0.03) &  0.15 (0.0008) &  0.28 (0.004) & 0.032 (0.005) \\ \bottomrule
		\end{tabular}
	}
	\caption{\rev{Summary of selected-feature ratios, reported as the fraction of regression coefficients that are nonzero or selected. Smaller values indicate sparser fitted models; the bold number indicates the sparsest result for the given taxonomic rank. Values in parentheses are standard deviations over $5$ independent runs. BayesMTL provides sparse multitask feature selection and identifies shared disease-category signals, while STL-LC can be sparser at several ranks because it fits independent single-task models.}}
	\label{tab: microbiome sparsity ratios}
\end{table}

\begin{figure}[htbp]
	\centering
	\begin{subfigure}{0.45\columnwidth}
		\centering
		\includegraphics[width=\textwidth]{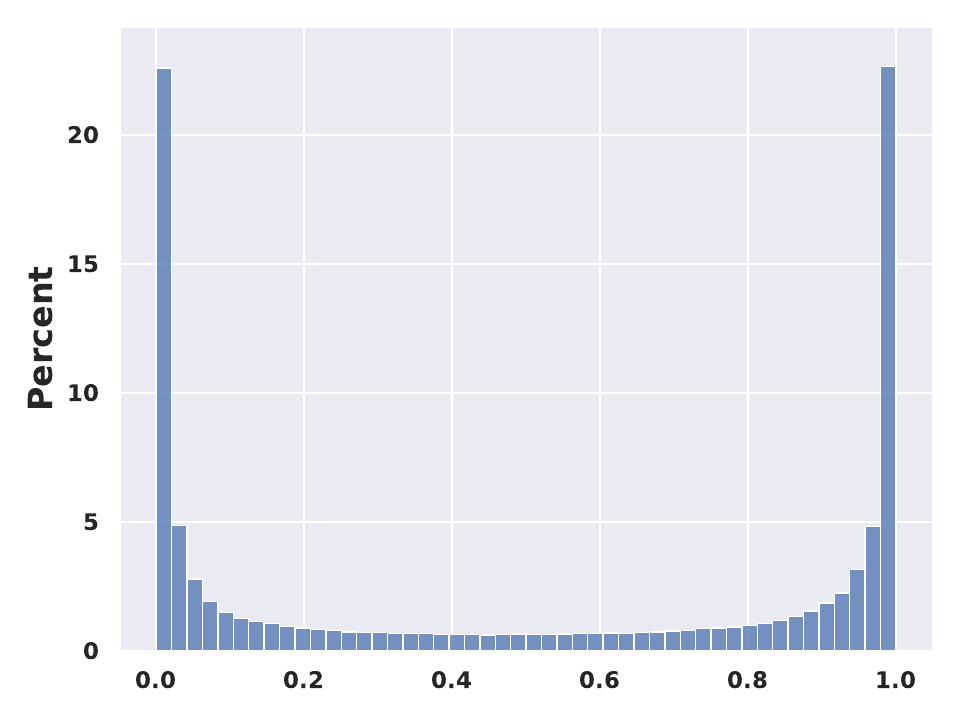}
		\caption{Predicted probabilities on training data.}
		\label{subfig: y_pred_train predict}
	\end{subfigure} 
	\begin{subfigure}{0.45\columnwidth}
		\centering		\includegraphics[width=\textwidth]{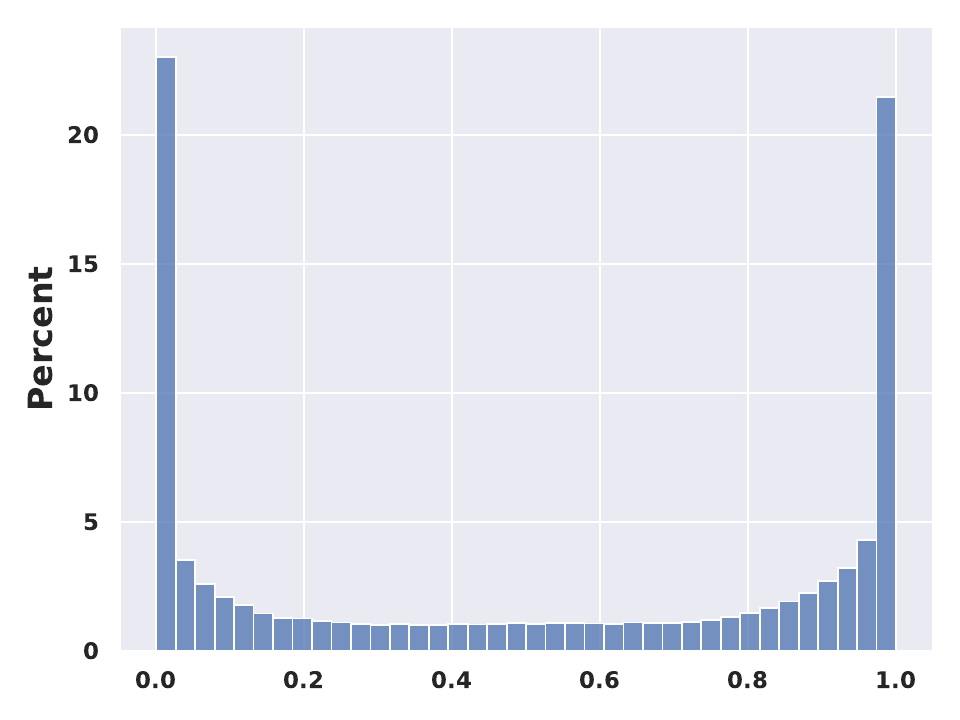}
		\caption{Predicted probabilities on test data.}
		\label{subfig: y_pred_test predict}
	\end{subfigure}
	
	\begin{subfigure}{0.5\columnwidth}
		\centering
		\includegraphics[width=\textwidth]{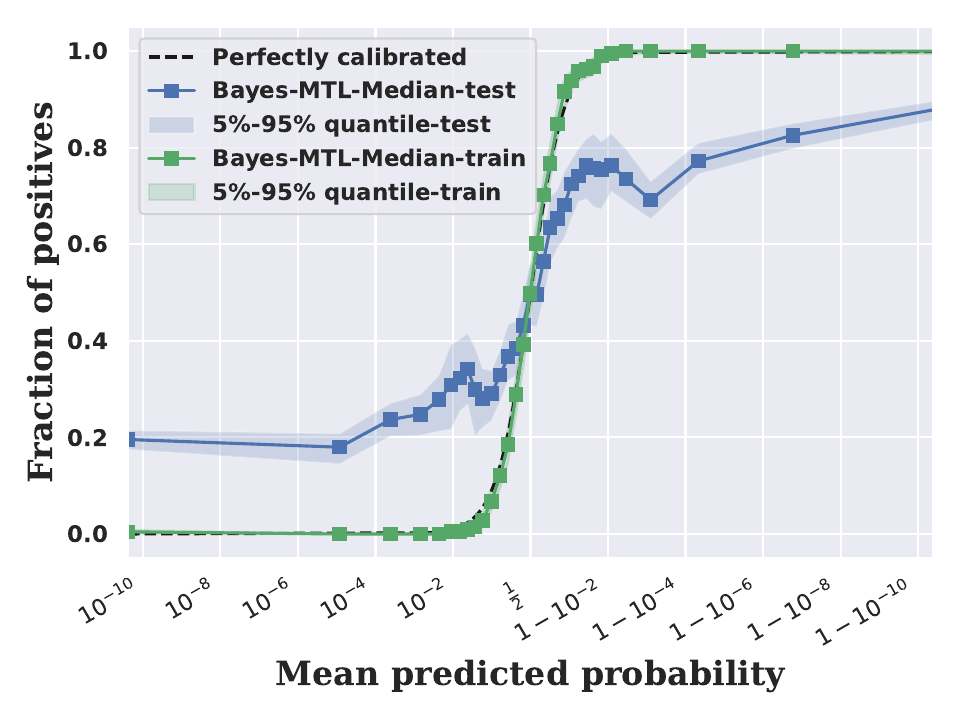}
		\caption{Calibration curves for the training data and the test data with 90\% marginal credible intervals.}
		\label{subfig: calibration}
	\end{subfigure}
	\caption{Calibration analysis for the proposed model on the Class taxonomic rank. Fig. (a) and Fig. (b) show the histograms of the predicted probabilities on the training and test data, respectively. Due to the choice of logit as link function, the predicted probabilities are concentrated around the boundaries. Fig. (c) shows the calibration curves of the predictions from training and test data. \rev{Training calibration is close to the diagonal over much of the probability range, while degradation on the test data near boundary values indicates overconfident predictions at the extremes.}}
	\label{fig: microbiome calibration plots}
\end{figure}

\rev{Calibration analysis complements the balanced-accuracy results by assessing how well the model's predicted probabilities reflect observed outcomes. Figure \ref{fig: microbiome calibration plots} shows close calibration on the training data across much of the probability range but weaker calibration on the test data, especially near the probability extremes. We therefore treat calibration as a diagnostic rather than as evidence of uniformly calibrated test-set probabilities. The overconfidence at the boundaries may arise from the logit link and from domain shift introduced by pooling heterogeneous microbiome studies. Even when predictive accuracy is mixed, the Bayesian formulation remains valuable because it exposes posterior uncertainty rather than returning only point predictions. Calibration curves for all taxonomic ranks are shown in the Supplementary Information document.}

\begin{figure*}[htbp]
	\centering
	\begin{subfigure}[b]{0.45\textwidth}
		\centering
		\includegraphics[width=\textwidth]{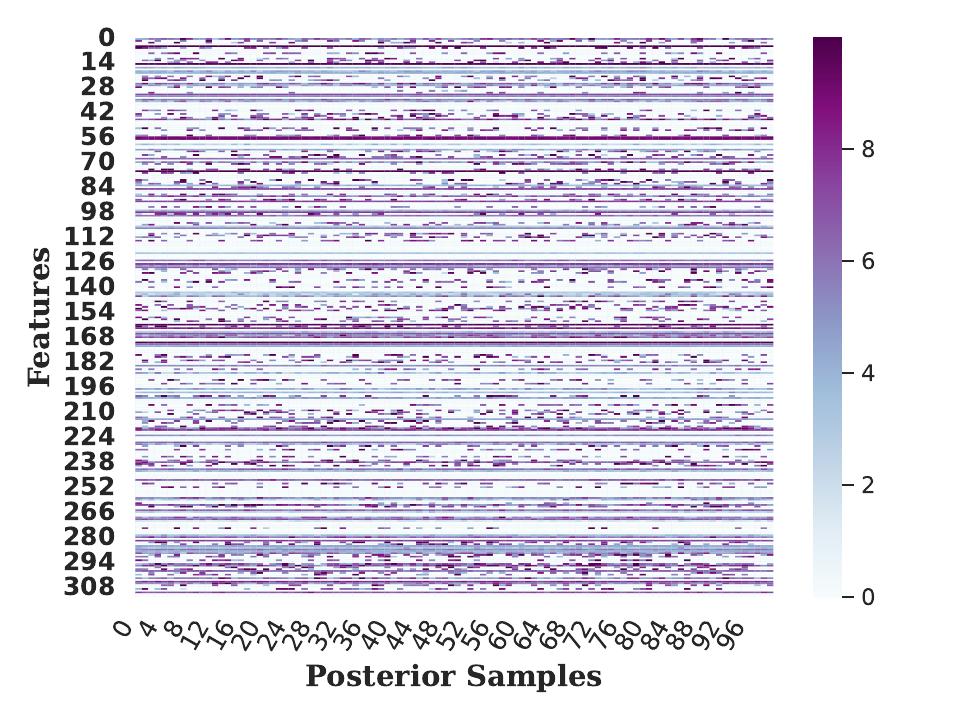}
		\caption{\footnotesize{Autoimmune Disease}}
	\end{subfigure}
	\hfill
	\begin{subfigure}[b]{0.45\textwidth}
		\centering
		\includegraphics[width=\textwidth]{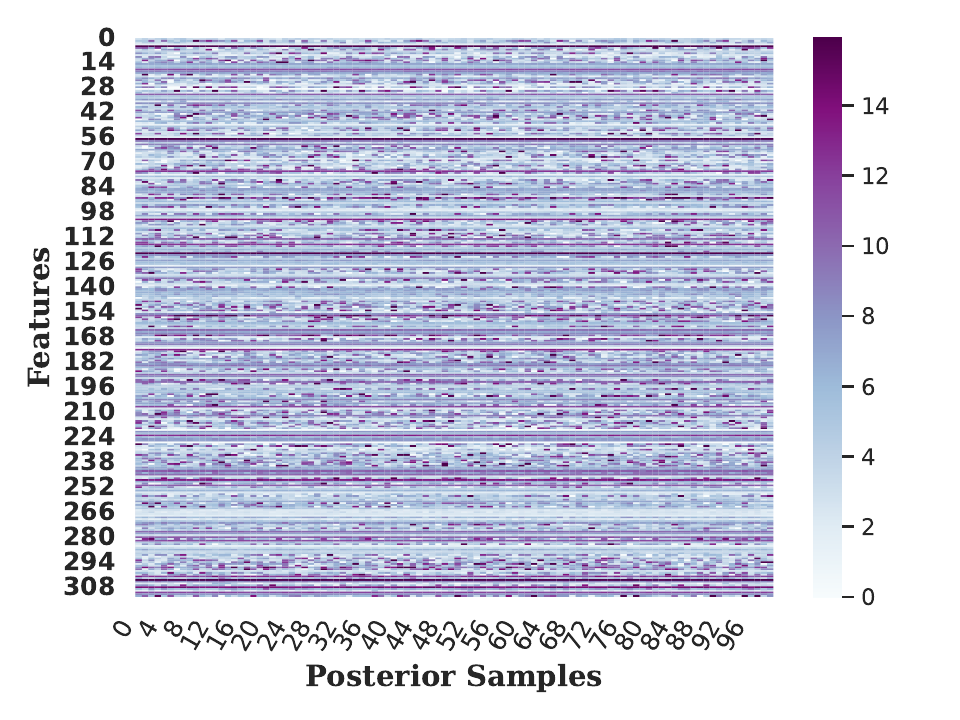}
		\caption{\footnotesize{Cancer}}
	\end{subfigure}
	
	\vspace{0.5em}
	
	\begin{subfigure}[b]{0.45\textwidth}
		\centering
		\includegraphics[width=\textwidth]{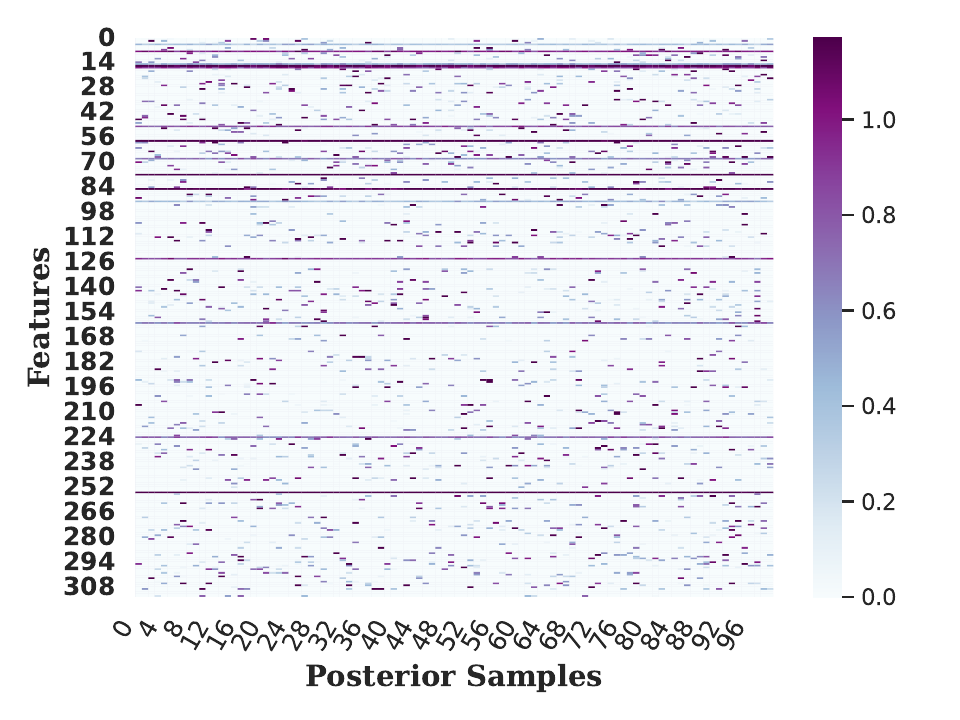}
		\caption{\footnotesize{\rev{Inflammatory Bowel Disease}}}
	\end{subfigure}
	\hfill
	\begin{subfigure}[b]{0.45\textwidth}
		\centering
		\includegraphics[width=\textwidth]{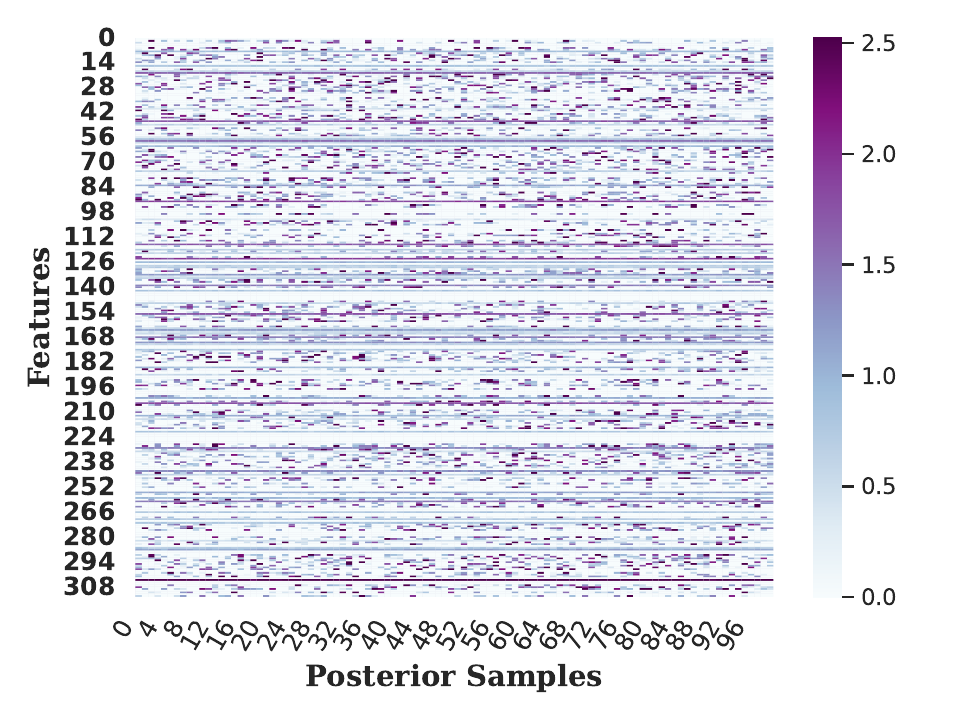}
		\caption{\footnotesize{Diarrhea}}
	\end{subfigure}
	
	\caption{Representative feature-sparsity patterns at the Class taxonomic rank. These four disease categories were selected because each pools multiple studies with non-trivial total sample sizes and together they illustrate sparse, intermediate, and dense feature-selection regimes. \rev{Inflammatory bowel disease and diarrhea show relatively small sets of repeatedly selected taxa}, autoimmune disease shows an intermediate pattern, and cancer shows broader support across taxa. The remaining disease categories for the Class taxonomic rank are provided in the Supplementary Information document.}
	\label{fig:microbiome_sparsity_posterior_class}
\end{figure*}

To interpret the fitted model, we combine posterior inclusion behavior and coefficient magnitude into a sparsity coefficient for each taxID. \rev{In the variational posterior, $\phi_j=q(z_j=1)$ is the posterior inclusion probability for feature $j$, $m_{tj}=\mathbb E_q(w_{tj})$ is the posterior mean task-specific coefficient, and $\phi_jm_{tj}$ provides a task-specific signed importance summary.} Figure \ref{fig:microbiome_sparsity_posterior_class} highlights four representative disease categories at the Class taxonomic rank. These examples were selected because they each combine multiple studies with non-trivial total sample sizes and, taken together, span sparse, intermediate, and dense feature-selection regimes. \rev{Inflammatory bowel disease and diarrhea are associated with relatively small sets of repeatedly selected taxa, autoimmune disease shows an intermediate pattern, and cancer yields a much broader set of informative features.} Rather than suggesting a universal microbiome signature, these results indicate that the degree of sparsity is itself disease dependent. This reinforces the value of a model that can represent both shared structure and uncertainty in feature selection. The remaining disease categories for the Class taxonomic rank, along with selection-stability curves across all taxonomic ranks, are provided in the Supplementary Information document. These allow comparison of sparsity levels across disease groups and taxonomic ranks.

\subsubsection{Diarrhea case study: feature importance}
To further examine the feature selection capability of the model, we focus on the diarrhea datasets \cite{david2015gut, kieser2018bangladeshi,mcdonald2018american} since they have shown a relatively consistent sparse patterns (Fig. \ref{fig:microbiome_sparsity_posterior_class}) and are directly relevant to gut and fecal microbiome studies. CLR transform is a normalization technique applied to each sample across features, therefore the magnitude of the transformed data varies across features. \rev{For global interpretation, $\phi_j$ summarizes posterior support for feature $j$, while $m_{tj}$ and $\phi_jm_{tj}$ summarize task-specific effect size and signed importance. For sample-level interpretation, we additionally include the observed CLR-transformed abundance.} We propose to assign feature importance of $j$-th feature (taxID) for task $t$ and sample $i$ by combining the regression coefficient ($\{ \mb w_{t} \}$), the sparsity parameters ($\mb z$) and the CLR transformed abundance data ($\mb x_t^i$):
\begin{equation}
	\label{eqn: feature importance} 
	\abs{\frac{w_{t,j} z_j x_t^{ij} }{\sqrt{\sum_j \left(w_{t,j} z_j x_t^{ij}\right)^2}}}.
\end{equation}
\rev{The proposed importance weight can be interpreted as the relative contribution to the predicted log-odds.} From the estimated posterior distribution, we draw samples to explore the full distribution of the feature importance. The most important features will correspond to the features with consistently high importance weights across draws, samples, and tasks. After obtaining important features for each taxonomic rank, we visualize the result using taxonomic lineage information, where lineages with at least $5$ taxIDs selected are shown in Fig. \ref{fig: microbiome lineage}. 

\subsubsection{Diarrhea case study: biological interpretation}
\label{subsubsec: diarrhea biology}
\begin{figure}[ht] 
	\centering
	\begin{subfigure}{0.62\textwidth}
		\centering
		\includegraphics[width=\textwidth]{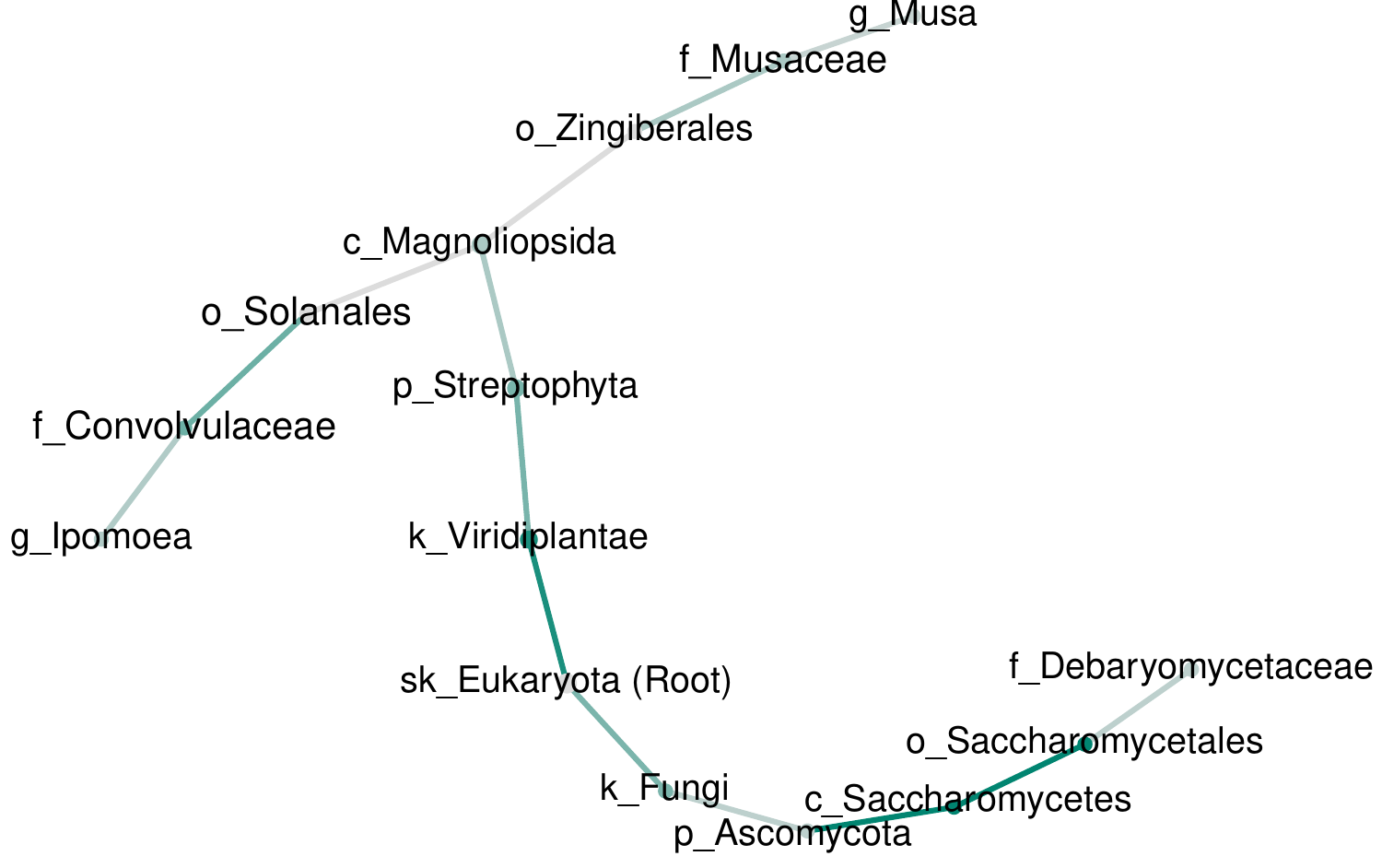}
		\caption{Eukaryota}
		\label{subfig: linage_Eukaryota}
	\end{subfigure}
	\begin{subfigure}{0.62\textwidth}
		\centering
		\includegraphics[width=\textwidth]{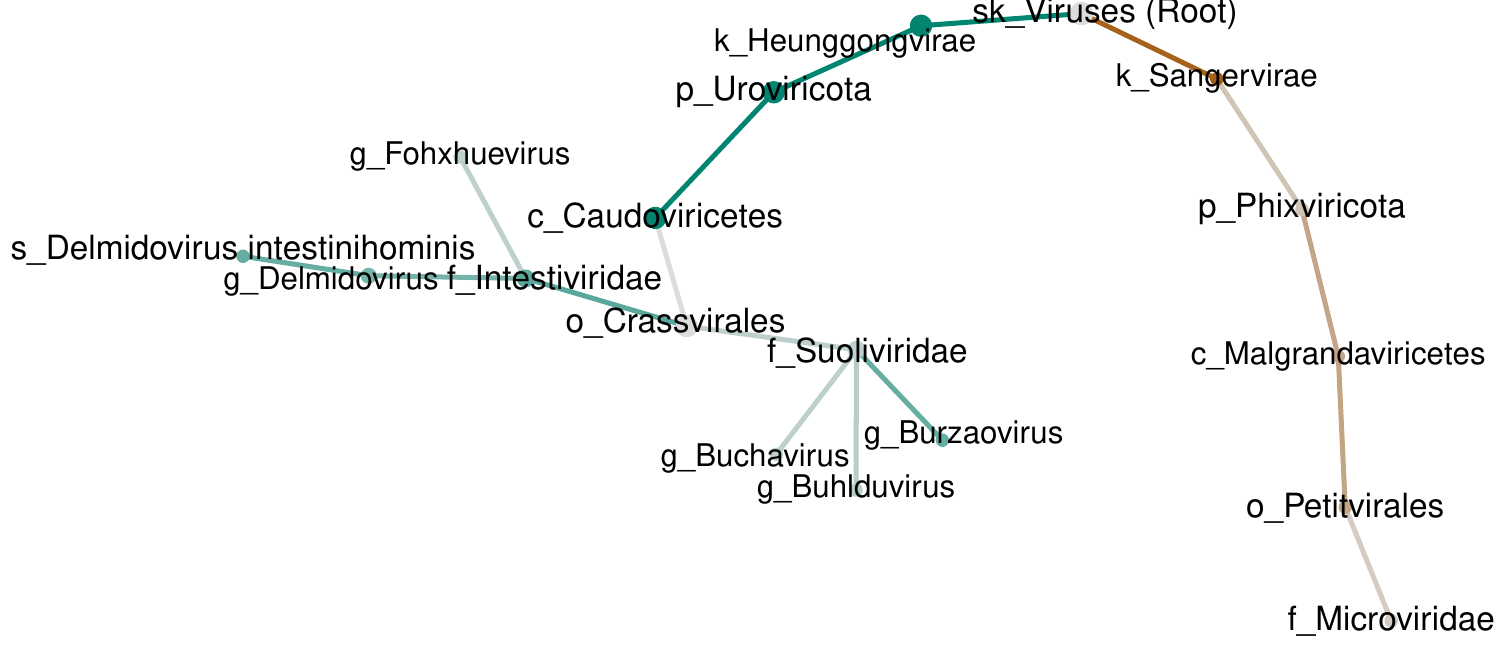}
		\caption{Viruses}
		\label{subfig: linage_Viruses}
	\end{subfigure}
	
	\begin{subfigure}{0.62\textwidth}
		\centering       
		\includegraphics[width=\textwidth]{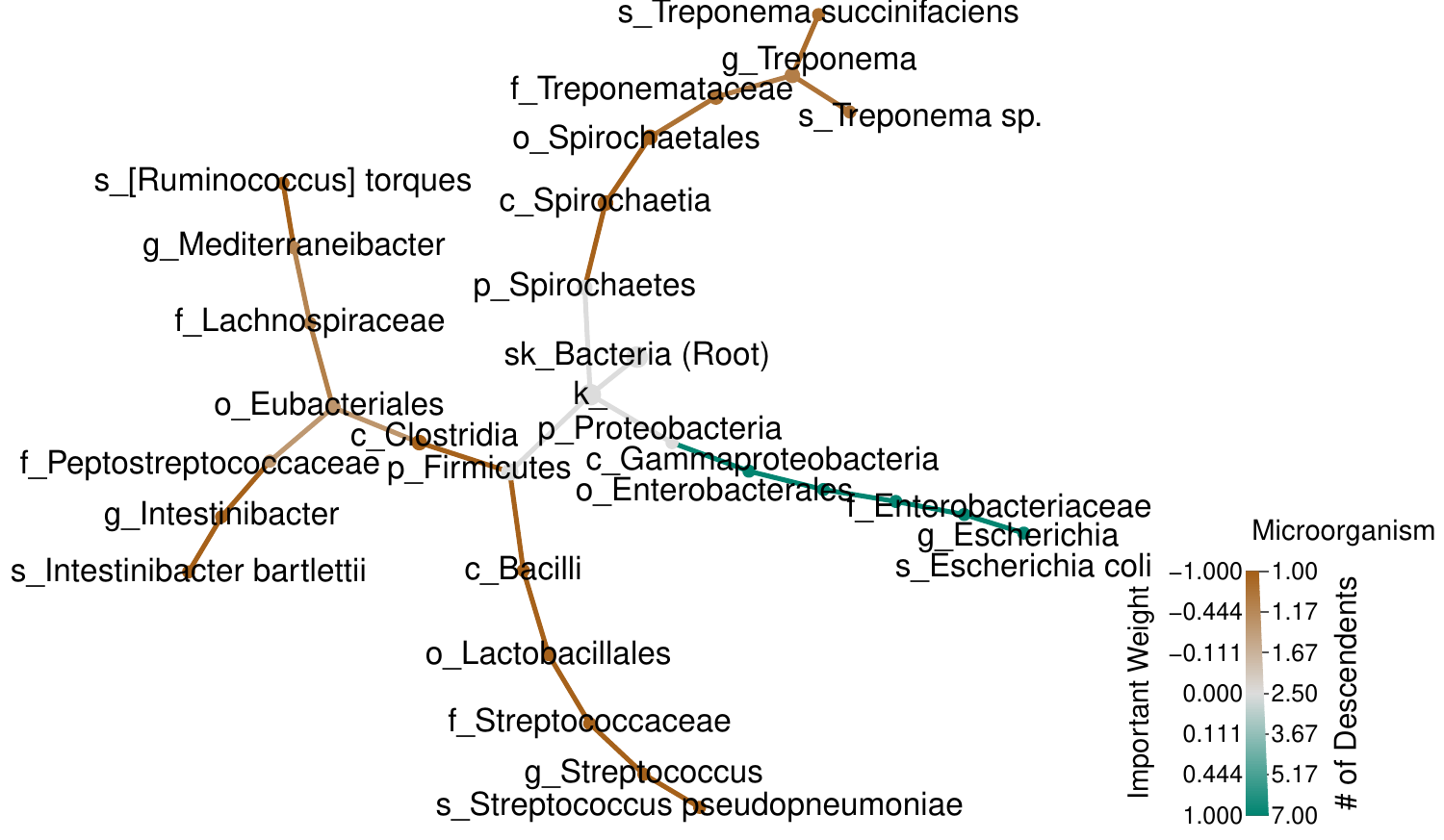}
		\caption{Bacteria}
		\label{subfig: linage_Bacteria}
	\end{subfigure}
	\caption{Lineage analysis of the important features in diarrhea datasets for three distinct taxonomic categories: Eukaryota, Viruses and Bacteria. Each node in the tree corresponds to a taxID. A positive sign (red colored) implies the given taxID is disease associated while a negative sign (green colored) implies the taxID is control associated. The identified lineages include \emph{Escherichia coli} and \emph{Crassvirales} bacteriophages, which have been studied in the context of diarrhea \cite{robins2002escherichia,okhuysen2010enteroaggregative,petro2020genetic,smith2023bacteriophages}.}
	\label{fig: microbiome lineage}
\end{figure}

\rev{The selected taxa in this case study are interpreted as hypothesis-generating associations rather than causal validations, because the pooled diarrhea cohorts differ in geography, population composition, and technical context.} The majority of lineages identified by the model are relevant in the context of the gut microbiome and diet, and highlights the association of \emph{Escherichia coli}, gut bacteriophages (order \emph{Crassvirales}) and a gut fungi, \emph{Debaryomycetaceae} with diarrhea. \emph{Enteroaggregative E. coli}, in particular has been routinely discussed to play a causal role in diarrhea \cite{robins2002escherichia,okhuysen2010enteroaggregative,petro2020genetic}. 
Although our analysis does link \emph{Petitvirales} to healthy controls, these results associate \emph{Crassvirales} with diarrhea (Fig. \ref{fig: microbiome lineage}), while previous reports link the abundance of \emph{Crassvirales} with gut health. \rev{We therefore interpret the \emph{Crassvirales} signal as a context-dependent, hypothesis-generating association that may reflect cohort composition, virome database incompleteness, or study-specific effects rather than a definitive disease association.} A number of factors may influence the gut microbiome composition including age, diet, and geographic location \cite{smith2023bacteriophages,gregory2020gut,ezzatpour2023human} thus the datasets used to establish this model may have been influenced by population demographics. Specifically, the diarrhea datasets used here are derived from a mixed population of subjects from Bangladesh and the United States \cite{david2015gut,kieser2018bangladeshi,mcdonald2018american}. Additionally, this analysis does not capture other global variables that can contribute to a disease state such as differences in virome richness and diversity \cite{zhu2018visualization}. Finally, differences in bioinformatics pipelines and reference databases used to classify sequences likely contributes to reported differences. For example, up to 99\% of phage sequences do not map to reference viral genomes \cite{gregory2020gut,ezzatpour2023human} thus increasing the likelihood of variability in identification of bacteriophage genomes, which are known for their complex and mosaic genome composition. Furthermore, the identification of plant-related features within the Magnoliopsida class suggest the capability of our approach to identify potentially disease-associated dietary components for hypothesis generation. Bacteriophages belonging to orders \emph{Crassvirales} and \emph{Petitvirales} are highly abundant in the human gut microbiome and have been associated with health \cite{ezzatpour2023human,shkoporov2019human,shkoporov2024dynamic,smith2023bacteriophages}. \revii{That a support shared across studies of the same disease converges on taxa independently reported in the context of diarrhea, such as \emph{Escherichia coli} and bacteriophages of the order \emph{Crassvirales}, indicates that the multitask structure captures reproducible signal rather than study-specific noise, and is therefore more likely to generalize across cohorts. Together with the posterior inclusion probabilities and task-specific coefficients, this yields an interpretable, uncertainty-aware shortlist of candidate taxa for downstream experimental validation, and is the sense in which the joint modeling produces informative findings beyond those obtained from separate single-task models.}

\section{Conclusions}
\label{sec: bmtl Conclusion}

In this work, we proposed a hierarchical sparse Bayesian multitask logistic regression model for predicting human health status from microbiome abundance data. The model leverages a shared sparsity structure to identify informative taxa across related tasks, while providing uncertainty quantification through a fully probabilistic framework. We developed an efficient variational inference procedure to enable scalable learning in high-dimensional settings.

Simulation studies demonstrate that the proposed approach performs particularly well when tasks share an underlying sparse structure, leading to improved feature recovery and predictive performance. \rev{On a real-world microbiome dataset pooled from multiple studies, the model achieves competitive predictive accuracy rather than uniform superiority over all single-task baselines, while consistently identifying sparse and interpretable sets of taxa. These results highlight its ability to extract biologically meaningful, uncertainty-aware signals from heterogeneous data while exposing calibration limitations at probability extremes.} \revii{The primary practical value of the multitask formulation on the real data is therefore interpretable, data-driven feature identification: it yields sparse, cross-study-consistent, uncertainty-aware candidate taxa for downstream experimental validation and hypothesis generation, a capability that is not available from independent single-task fits and that is more likely to transfer across cohorts.}

There are several directions for future work. One natural extension is to explore alternative link functions, such as the probit link, which have flatter tails and may reduce the overconfident predictions observed with the logit formulation. Another direction is to incorporate additional hierarchical structure into the model to account for relevant metadata, such as study-specific factors or patient demographics. \rev{The current model does not explicitly correct residual batch effects or include study-level technical covariates, demographic variables, body-site indicators, or geography as covariates; incorporating such terms, hierarchical random effects, or explicit batch-adjustment components is an important future direction. A more exhaustive prior-sensitivity analysis is also left for future work.}

It is also of interest to extend the model to more general prediction settings. In particular, a multi-label formulation would allow each task to include multiple binary outcomes, reflecting the fact that diseases are not mutually exclusive in many clinical settings. A related extension is to consider multiclass classification, where each task involves more than two outcome categories, such as different stages of disease progression.

\backmatter

\bmhead{Supplementary information}

A supplementary information document accompanies this manuscript, containing references and metadata for all microbiome studies used in this work, as well as additional results.

\bmhead{Acknowledgements}

Not applicable.

\section*{Declarations}

\textbf{Funding}

This work was performed under the auspices of the U.S. Department of Energy by Lawrence Livermore National Laboratory under Contract DE-AC52-07NA27344 (LLNL-MI-853606). It was supported by the LLNL Laboratory Directed Research and Development Program, the ETI Consortium under US Department of Energy grant DE-NA0003921, and the US Army Research Office under grant number W911NF1910269. 

\textbf{Competing interests}

The authors declare that they have no competing interests.

\textbf{Ethics approval and consent to participate}

This study used previously published, de-identified metagenomic datasets obtained from public repositories. No new human participants were recruited and no directly identifiable human data were collected by the authors for this study.

\textbf{Consent for publication}

All authors have read and approved the manuscript for publication.

\textbf{Availability of data and materials}

The datasets utilized in this study can be found in \url{https://zenodo.org/records/17315984}.

\textbf{Code availability}

The code that implements the Bayesian MTL model proposed in this paper can be found here: https://github.com/llnl/BayesMTL

\textbf{Authors' contributions}

HZ: Conceptualization, Investigation, Methodology, Computational Experiments, Formal analysis, Software, Writing – original draft.
ARG: Conceptualization, Methodology, Software, Writing – review \& editing.
CRK: Data curation, Formal analysis, Writing – review \& editing. 
CV: Methodology, Writing – review \& editing.
HR: Methodology, Writing – review \& editing. 
BZ: Methodology, Writing – review \& editing. 
JMM: Data curation, Writing – review \& editing. 
JBT: Data curation, Writing – review \& editing. 
NJM: Data curation, Writing – review \& editing. 
MB: Methodology, Writing – review \& editing. 
CJ: Project administration, Funding acquisition, Writing – review \& editing. 
AH: Methodology, Supervision, Discussion, Writing – review \& editing.
NAB: Project administration, Funding acquisition, Supervision, Writing – review \& editing.

\bibliography{references,additional_ref,datasets}

\end{document}